\documentclass[final]{elsarticle}

\usepackage[utf8]{inputenc}
\usepackage{amsmath}
\usepackage{amsfonts}
\usepackage{amssymb}
\usepackage{graphicx}
\usepackage{fullpage}
\usepackage[pdftex]{hyperref}
\usepackage{color}
\usepackage[titletoc,title]{appendix}
\usepackage[mathscr]{eucal}
\usepackage[numbers]{natbib}
\usepackage{floatrow}
\newfloatcommand{capbtabbox}{table}[][\FBwidth]

\newtheorem{thm}{Theorem}[section]

\newtheorem{prop}[thm]{Proposition}
\newtheorem{corol}[thm]{Corollary}
\newtheorem{defn}{Definition}[section]
\newtheorem{remark}{Remark}[section]

\newcommand{\proof}{\noindent {\bf Proof.}~}

\DeclareMathOperator*{\diag}{diag}

\newcommand{\model}{ES$^2$N}
\newcommand{\linear}{linearSCR}

\begin{document}

\title{Edge of Stability Echo State Network}
%:
%Orthogonal Additive Echo State Network: 
%a nonlinear reservoir computing model immune to the degradation of memory
%Nonlinear Reservoir Computing %without Memory Degradation.
%with Long Short-Term Memory.
%}

\author[1]{Andrea Ceni}\ead{andrea.ceni@di.unipi.it}\corref{cor1}

\author[1]{Claudio Gallicchio}\ead{claudio.gallicchio@unipi.it}

\cortext[cor1]{Corresponding author}
\address[1]{Department of Computer Science, University of Pisa, Largo Bruno Pontecorvo, 3 - 56127, IT}

\begin{abstract}
Echo State Networks (ESNs) are time-series processing models working under the Echo State Property (ESP) principle. The ESP is a notion of stability that imposes an asymptotic fading of the memory of the input. 
On the other hand, the resulting inherent architectural bias of ESNs may lead
%too much of fading memory leads 
to an excessive loss of information, which in turn harms the performance %of ESNs 
in certain tasks with long short-term memory requirements.
With the goal of bringing together the fading memory property and the ability to retain as much memory as possible, in this paper we introduce a 
%To fulfil the need of fading memory systems able to retain as much memory as possible, we propose a 
new ESN architecture, called the Edge of Stability Echo State Network (\model).
The introduced \model{} model is based on defining the reservoir layer as a convex combination of a nonlinear reservoir (as in the standard ESN), and a linear reservoir that implements an orthogonal transformation.
We provide a thorough mathematical analysis of the introduced model, proving that the whole eigenspectrum of the Jacobian of the \model{} map can be contained in an annular neighbourhood of a complex circle of controllable radius, and exploit this property to demonstrate that the \model's forward dynamics evolves close to the edge-of-chaos regime by design.
Remarkably, our experimental analysis shows that the newly introduced reservoir model is able to reach the theoretical maximum short-term memory capacity. At the same time, in comparison to standard ESN, \model{} is shown to offer an excellent trade-off between memory and nonlinearity, as well as a significant improvement of performance in autoregressive nonlinear modeling.
\end{abstract}

\maketitle

%\newpage

%\tableofcontents

%\newpage

\section{Introduction} 
\label{sec:introduction}

Recurrent neural networks (RNNs) \cite{kolen2001field} are computational models designed to extract features from data with temporal structures. Applications ranges from speech recognition to classification of time series. The most common way of training RNNs is via stochastic gradient descent methods, usually via the backpropagation through-time algorithm \cite{werbos1990backpropagation}. Unfortunately, these methods come with a significant computational effort. Modern hardware unleashed the power of parallel computing techniques, allowing to reduce the computational time of training deep learning models as RNNs. However, the price to pay is a massive energy consumption.
Moreover, a fundamental limitation of theoretical nature prevents RNNs to be fully exploited, namely the vanishing/exploding (V/E) gradient issue \cite{bengio1994learning}.
In this regard, an appealing alternative is represented by Reservoir Computing (RC) \cite{nakajima2021reservoir,lukovsevivcius2009reservoir}, a different paradigm of training RNNs dodging the V/E while being computationally fast, and energy efficient.
% #### NEW Refs #################
The flexibility of the RC paradigm offers a suitable theoretical framework for computing with physical substrates \cite{liang2022rotating,tan2023dynamic,wu2023wearable,wang2023echo}, for fast and scalable graph neural networks models \cite{gallicchio2020fast,cini2023scalable}, and for implementing digital twins of real-world nonlinear dynamical systems \cite{kong2023reservoir}. 
% ###############################
The key idea of RC is to inject the input signal into a large random untrained recurrent layer, the \emph{reservoir}, from which a readout layer is optimised to fit the desired target signal. 
The Echo State Network (ESN) \cite{jaeger2001echo, jaeger2004harnessing} provides a popular discrete-time class of RC machines. The name ESN recalls the imagery of the input signal echoing and reverberating within the pool of neuronal activations in the reservoir, which in turn serves as an high-dimensional representation of the past history of the input. 
Although gradients are not backpropagated in the RC paradigm, the ESN's forward dynamics are ruled by the very same equation of conventional RNNs. 
Thus, ESNs inherit a problem which closely relates to the V/E gradient problem of plain RNNs, namely the degradation of memory \cite{inubushi2017reservoir}.
%############
As revealed by previous studies \cite{dambre2012information,schulte2023refined}, the degradation of memory is linked to the nonlinearity of the system in an inherent trade-off.
Nonlinear computation and short-term memory are two fundamental aspects of neural systems. Therefore, the existence of a trade-off between them compels to design nonlinear RC systems able to retain as much memory as possible.
In fact, the memory capacity is a key feature to reach desirable results in certain learning tasks \cite{carroll2022optimizing}.

\iffalse
%###########################
Nonlinear computation and short-term memory are two fundamental aspects of neural systems. 
Previous studies revealed that the degradation of memory is linked to the nonlinearity of the system in an inherent trade-off, which makes compelled the research of Reservoir Computing (RC) architectures that tackle this problem.
%###########################
Interestingly, the nonlinearity and the short-term memory of the system appear to contrast each other as a (potentially) universal property that holds for any dynamical system \cite{dambre2012information,schulte2023refined}.
The identification of this trade-off renders interesting to design nonlinear RC systems able to retain as much memory as possible.
In fact, the memory capacity is a key feature to reach desirable results in certain learning tasks \cite{carroll2022optimizing}.
%###########################
\fi

In this paper, we propose and analyse a novel RC model, called \emph{Edge of Stability Echo State Networks (\model)}, that copes with the degradation of memory of nonlinear reservoir systems.
We tackle the degradation of memory of RNNs from a dynamical system perspective, framing the problem within the \emph{edge of chaos} hypothesis.
The proposal of this hypothesis can be traced back at least to the late eighties \cite{langton1990computation,packard1988adaptation}, where it has been observed that extensive computational capabilities are achieved by adaptive systems whose dynamics are neither chaotic nor ordered but somewhere in between order and chaos.
In this paper we link these ideas rooted in the study of adaptive systems to the case of partially randomised RNNs, as ESNs are.

ESNs work “properly" provided with the so called Echo State Property (ESP) \cite{jaeger2001echo,yildiz2012re}. 
In coarse terms, the ESP guarantees the ESN to possess a unique input-driven solution such that all the trajectories originating from different initial conditions (in the infinite past) synchronise with it (in present time). Following the imagery, such a unique input-driven solution would represent the echo of the input signal from the infinite past.
The simplest known criterion to ensure the ESP is to impose the maximum singular value of the reservoir's connections matrix to be less than one. This condition implies straight contraction in the phase space at each time step. Therefore, any two different internal states of the RNN, when driven by the same input sequence, will get closer and closer to each other as time flows ahead. Although stable, such a dynamical system would have truly little margin to exploit the transient dynamics for computational purposes, due to the straightforward contraction.
An ideal situation would be for the RNN to stay in a regime of balance between stable contractive dynamics and unstable chaotic dynamics, i.e. along the \emph{edge of chaos} \cite{bertschinger2004real, legenstein2007makes}.
This led the RC community to adopt the rule of thumb of setting the reservoir matrix to have spectral radius approximately one.
However, this rule of thumb inevitably overweights the contractive dynamics, especially when considering the action of the input driving neurons towards the saturation regime.
This overweighting of the contractive dynamics reflects in the degradation of memory in the forward dynamics of ESNs. A stronger condition would be to have the entire eigenspectrum of the reservoir matrix to be “around" the complex unit circle.
We study the eigenspectrum of \model, and prove in Theorem~\ref{thm:annular} that all the eigenvalues of an \model{} lie within an annular neighbourhood of the complex unit circle. The radius of such neighbourhood can be tuned via a specific hyperparameter of the \model{} that we called the \emph{proximity} hyperparameter.
Moreover, in the limit of small values of the proximity hyperparameter, we prove in Theorem~\ref{prop:edge_chaos} that the \model{} dynamics narrowly hover over the edge of chaos while being on average over time in the stable regime, hence the name Edge of Stability Echo State Network.
Through experiments, we show the empirical advantages of the proposed approach in terms of short-term memory capacity, memory-nonlinearity tradeoff, and autoregressive time-series modeling.\\
The rest of this paper is organized as follows. In Section \ref{sec:ESN} we introduce the reader to the RC fundamentals and the classical %leaky 
ESN model. 
In Section \ref{sec:ESESN} we propose our \model{} model and provide a theoretical analysis of its dynamics. Section \ref{sec:experiments} is dedicated to the experimental results, focusing on the short-term memory capacity, the trade-off between nonlinearity and memory, and autoregressive time series modeling.
Finally, in Section~\ref{sec:conclusions}, we discuss our findings, draw the conclusions, and point out interesting research directions to explore further.

\section{The ESN model} 
\label{sec:ESN}
RC neural networks \cite{cucchi2022hands,verstraeten2007experimental} identify a class of fastly trainable RNNs, in which a non-linear untrained dynamical layer is followed by a linear trainable readout component. In this contribution, we focus on the Echo State Networks (ESNs) \cite{jaeger2004harnessing,jaeger2001echo} approach within RC, and we recall the 
%First, we recall the 
well-established \cite{jaeger2007optimization, lukovsevivcius2009reservoir, lukovsevivcius2012practical} formulation of the leaky ESN model, 
%with linear readout, 
which consists of a nonlinear reservoir layer made up of leaky recurrent neurons followed by a linear readout.
The equations read as follows:
%The leaky-ESN with linear readout reads
\begin{align}
\label{eq:leaky_esn}
x[t] =& \alpha \tanh(\rho\mathbf{W}_{r} x[t-1] + \omega\mathbf{W}_{in}u[t]) + (1-\alpha) x[t-1], \\
\label{eq:esn_output}  
z[t] =& \mathbf{W}_{o} x[t].
\end{align}
The internal state $x[t] $, input $u[t]$, and output $z[t]$ are, respectively, $N_r$-dimensional, $N_i$-dimensional and $N_o$-dimensional vectors of real values. 
The nonlinearity is expressed by the element-wise applied hyperbolic tangent function $\tanh$, and the system is typically initialised in the origin, i.e., $x[0] = 0$.
%The reservoir $\mathbf{W}_{r}$, input-to-reservoir $\mathbf{W}_{in}$, and readout $\mathbf{W}_{o}$ matrices have dimensions, respectively, $N_r \times N_r $ , $ N_r \times N_i $, and $ N_o \times N_r $.
Matrices $\mathbf{W}_{r}, \mathbf{W}_{in}$ are, respectively, the recurrent reservoir weight matrix and the input weight matrix, both randomly instantiated and left untouched. 
In this paper, we initialise $ \mathbf{W}_{in} $ with i.i.d. random uniformly distributed entries in $(-1,1)$, and $ \mathbf{W}_{r} $ with i.i.d. normally distributed entries with zero mean and standard deviation $\dfrac{1}{\sqrt{N_r}}$.
This initialisation scheme for $ \mathbf{W}_{r} $ ensures that, for large $N_r$, the spectral radius of $ \mathbf{W}_{r} $ is approximately 1, thanks to the circular law from random matrix theory \cite{meckes2021eigenvalues}.
%Rooted in the circular law from random matrix theory, this initialisation scheme for $ \mathbf{W}_{r} $ ensures, in the limit of an infinitely large reservoir, that the spectral radius of $ \mathbf{W}_{r} $ is 1. 
%The resulting term $\rho \mathbf{W}_{r}$ has a spectral radius of $\rho$.
This allows us to interpret the hyperparameter $\rho$ as the spectral radius (i.e., the largest eigenvalue in modulus) of the effective recurrent matrix, i.e. of $ \rho \mathbf{W}_{r} $.
 %of $ \mathbf{W}_{r} $.
While, the input scaling $\omega$ is an hyperparameter entitled to rescale the weight of the current input into the reservoir dynamics. 
The leaky ESN owes its name to the presence of the hyperparameter $\alpha \in (0,1]$, the \emph{leak rate}, that is designated to modify the time scale of the ESN dynamics according to the input at hand.
In this paper we always consider the recurrent reservoir matrix $\mathbf{W}_{r}$ a fully connected weight matrix.

\subsection{Training ESNs via ridge regression}

Given a training set of input-target samples, $\{u[t],y[t]\}_{t = 1, \ldots, T}$, we train a leaky ESN by means of optimising the readout matrix $\mathbf{W}_{o}$ in order to solve the linear regression problem $ y[t]=\mathbf{W}_{o}x[t] $.
Usually, this is achieved via \emph{ridge regression (or Tichonov regularisation)} \cite{lukovsevivcius2009reservoir} by means of the following formula:
\begin{equation}
\label{eq:ridge_regress}
    \mathbf{W}_{o} = \mathbf{Y} \mathbf{X}^T (\mathbf{X} \mathbf{X}^T + \mu \mathbf{I})^{-1},
\end{equation}
where $ \mathbf{X} $ is the matrix of dimension $ N_r \times T $, containing all the internal states $x[t]$ of the ESN driven by the input $u[t]$ for $k=1, \ldots, T$, $\mathbf{Y}$ the matrix of dimension $ N_o \times T $, containing all the target values $y[t]$, $ \mathbf{I} $ is the identity matrix of dimension $N_r \times N_r $, and $\mu$ is the regularisation parameter.
Throughout the paper we denote the transposed of a matrix $ \mathbf{X} $ with $ \mathbf{X}^T $.
Usually, when dealing with regression tasks, a number of initial time steps are discarded for the computation of eq.~\eqref{eq:ridge_regress} to allow the reservoir to “warm up". This is to ensure that the ESN transient dynamics are washed out, so that the internal dynamics of the ESN get linked with the driving input.

\subsection{Echo state property}
%In the reservoir state transition \eqref{eq:leaky_esn}, the input scaling $\omega$ is an hyperparameter entitled to rescale the weight of the current input into the reservoir dynamics. 
%The hyperparameter $\rho$ is a positive real value controlling the amount of nonlinearity into the reservoir and the contribution of the past activations.
ESNs work under the fundamental assumption of the ESP, a condition ensuring a unique stable input-driven response \cite{jaeger2001echo}. Roughly speaking, the ESP guarantees that the internal state $x[t]$ is uniquely determined by the entire past history of the input signal.
The easiest condition to ensure the ESP is to set a reservoir such that
\begin{equation}
    \label{eq:suff_cond}
    \lVert \rho\mathbf{W}_{r} \rVert < 1,
\end{equation}
where, for a given matrix $ \mathbf{M} \in \mathbb{R}^{N\times N}$, $\lVert \mathbf{M} \rVert  $ denotes the matrix norm induced by the Euclidean norm in $\mathbb{R}^N$, or equivalently the maximum singular value of $ 
\mathbf{M} $. For the rest of the paper, we will always use $ \lVert \cdot \rVert $ to denote either the Euclidean norm on $\mathbb{R}^N$, or the matrix norm induced by that, depending on whether the argument of the norm is respectively a vector or a matrix. \\ 
The condition expressed by eq.~\eqref{eq:suff_cond} implies contraction of the internal states at each time step for whatever input; that is a strong stability condition implying a Markovian state space organisation \cite{gallicchio2011architectural}. However, stability is not the only property we crave from an ideal recurrent neural system.
%Sufficient conditions to ensure the ESP have been established mainly focusing on the maximum singular value $\sigma$ of the reservoir matrix, namely imposing that $\sigma<1$, neglecting the role of the input \cite{buehner2006tighter, lukovsevivcius2009reservoir, yildiz2012re}.
We strive for a stable dynamical system that is able to recall the information conveyed from the external input for as long as possible.
In other words, we want to keep the ESP while staying close to the border of its domain of existence.
This led the RC community to adopt the less restrictive rule of thumb of setting the spectral radius of $  \rho\mathbf{W}_{r} $ approximately (or often slightly less than) $ 1 $, e.g. initialising $ \mathbf{W}_{r} $ to have spectral radius $1$ and then setting setting the hyperparameter $\rho\approx 1 $.
Intuitively, $\rho$ controls the amount of nonlinearity into the reservoir and the contribution of the past activations. A small $\rho$ promotes stable dynamics at the expenses of forgetting faster the past internal activations, thus amplifying the fading memory property.
However, as previous works demonstrated \cite{yildiz2012re}, the only constraint of $\rho<1$ is not generally sufficient to guarantee the ESP in an input-driven ESN; and, due to the presence of the external input driving the dynamics, it is not even necessary.
In fact, the ESP is not a property of the reservoir alone, but rather of the reservoir plus the forcing input.
Some efforts to comprehend the input-dependence in the analysis of the ESP in RC have been exerted in \cite{manjunath2013echo,wainrib2016local,gallicchio2018chasing,ceni2020echo}. As demonstrated in the literature, the ESP might hold even with large values of $\rho$, as long as the amplitude of the inputs (or the hyperparameter $\omega$) are large enough to counterbalance the effect. As a consequence, $\rho$ should be optimised in synergy with the input scaling $\omega$.
%From a dynamical system perspective, it would be ideal to find a sweet combination of $\rho$ and $ \omega $ such that the overall nonautonomous dynamics of the input-driven recurrent neural network hover over the edge between chaotic dynamics and stable contractive dynamics where the ESP holds.

%AGGIUNGERE POSSIBILMENTE QUALCHE OSSERVAZIONE DI COLLEGAMENTO CON IL RESTO. Ad esempio sul fato che sebbene le esn presentino dei vantaggi, restano complicate da tunare all edge of chaos a priori (insomma un link rispetto al fatto che comunque restano biasate dalla markovianita', mentre resta la necessita' di esplorare bias architetturali differenti se pur all interno della echo state property di una esn standard...link con quello che facciamo in questo paper)

Although ESNs present many advantages, it remains unclear how to tune a priori an ESN close to the edge of chaos keeping the benefits of a stable nonlinear computational system. Driven by the need to reconcile the properties of stability, nonlinearity, long-term memory, and ease of computation, we introduce in the next section a new RC architecture. 

\section{The \texorpdfstring{\model}{} model} 
\label{sec:ESESN}

We propose a variant of ESN, called Edge of Stability ESN (\model), whose equations with linear readout read as follows:
\begin{align}
\label{eq:state_update}
x[t] =& \beta  \; \phi(\rho\mathbf{W}_{r} x[t-1] +  \omega\mathbf{W}_{in}u[t]) + (1-\beta)\;  \mathbf{O} x[t-1], \\
\label{eq:es2sn_output}  
z[t] =& \mathbf{W}_{o} x[t],
\end{align}
Where $x[t]$, $u[t]$, $z[t]$, $\mathbf{W}_r$, $\mathbf{W}_{in}$, and $\mathbf{W}_o$ are as in eq.~\eqref{eq:leaky_esn} and eq.~\eqref{eq:esn_output}.
Here, in addition, the matrix $\mathbf{O}$ is a randomly generated \emph{orthogonal} matrix, and $\beta \in (0,1] $ a real valued hyperparameter that we call \emph{proximity}.
%############
All the random orthogonal matrices in the experiments of this paper have been obtained by generating first a random matrix $\mathbf{D}$ of the desired dimension $N_r\times N_r$ with i.i.d. uniformly random entries in $(-1,1)$, hence performing a QR decomposition of $\mathbf{D}=\mathbf{QR}$, and taking the resulting $N_r\times N_r$ orthogonal matrix $\mathbf{Q}$ as the random orthogonal matrix $\mathbf{O}$ in eq.~\eqref{eq:state_update}.
%###############
Although apparently the hyperparameters $\alpha$ of a leaky ESN of eq.~\eqref{eq:leaky_esn} and $\beta$ of an \model {} of eq.~\eqref{eq:state_update} share the same position in the equation, they play quite a different role in the dynamics of the RNNs, as it will be evident after our theoretical and experimental analysis. In fact, while the value of $\alpha$ in eq.~\eqref{eq:leaky_esn} is intended to slow down the speed of the reservoir dynamics relative to those of the input signal \cite{jaeger2007optimization}, as we will see later in this section, the role of $\beta$ in eq.~\eqref{eq:state_update} is to determine the proximity of the reservoir dynamics to the edge of chaos. % For this reason, we call $\beta$ the \emph{proximity} hyperparameter.
With the setting in eq.~\eqref{eq:state_update}, the reservoir of the \model {} results into a convex combination of a standard nonlinear input-driven reservoir (first term on the right hand side) and a linear orthogonal input-free reservoir (second term on the right hand side). 
Note that for $\beta \rightarrow 0 $, and $\alpha \rightarrow 0$, the \model{} model of eq.~\eqref{eq:state_update} and the leaky ESN model of eq.~\eqref{eq:leaky_esn} both coincide to the standard (non-leaky) ESN model.
%However, it is likely to obtain further improvements to performance by optimising separately $\beta_1,\beta_2$. 

%the dynamical reservoir properties for the newly introduced \model {} model, focusing on 
%In particular, in Section~\ref{sec:critical} we focus on analysing the resulting critical dynamical regime of the nonlinear reservoir, while in Section~\ref{sec:linear_ESESN} we further elucidate the model's properties by focusing on the analysis of its dynamics in the case of linear activation function.

\subsection{Edge of chaos in \texorpdfstring{\model}{}}
\label{sec:critical}

In this section we present a detailed mathematical analysis of the structure of the eigenspectrum for the newly introduced \model {} model, and its maximum local Lyapunov exponent. 
We start introducing our notation.
%From here on, we will drop the notation of the hyperparameters $\rho$ and $\omega$ to simplify the expressions, but without loss of generality.
Denoting the \model {} reservoir map as $G(u,x)= \beta \phi(\rho\mathbf{W}_{r} x +  \omega\mathbf{W}_{in}u) + (1-\beta) \mathbf{O} x$, where we use $\phi$ to generally indicate the nonlinear activation function, then 
the corresponding Jacobian reads as follows:
%prev:the Jacobian of the ESESN map reads 
\begin{equation}
    \label{eq:jacob_ESESN}
    \dfrac{\partial G}{\partial x} (u,x) = \beta \,\,  \mathbf{D}_{(u,x)}\,\,  \rho\mathbf{W}_{r}  \,\, + \,\, (1-\beta)\,\,  \mathbf{O},
\end{equation}
where we defined the following diagonal matrix 
\begin{equation}
    \label{eq:diag}
    \mathbf{D}_{(u,x)}  := \diag(\phi'(\rho\mathbf{W}_{r} x +  \omega\mathbf{W}_{in}u)) ,
\end{equation}
whose supremum of its norm we denote as follows
\begin{equation}
    \label{eq:gamma}
    \gamma := \sup_{u,x} \lVert \mathbf{D}_{(u,x)} \rVert .
\end{equation}
Note that whenever $ |\phi'|\leq 1$, then it holds that $\gamma \in [0,1]$.
Considering a specific input-driven trajectory $ \{ ( u[t+1], x[t] ) \}_{t=0, \ldots, T-1} $, we denote as
\begin{equation}
    \label{eq:inpdriven_jacob}
    \mathbf{J}[t] :=  \dfrac{\partial G}{\partial x} (u[t+1],x[t])
\end{equation}
the Jacobian map of eq.~\eqref{eq:jacob_ESESN} evaluated along the input-driven trajectory.
Finally, we will denote the maximum singular value of the matrix $ \rho \mathbf{W}_r $ as follows
\begin{equation}
    \label{eq:sigma_reservoir}
    \sigma :=  \lVert \rho\mathbf{W}_{r} \rVert .
\end{equation}

We start our analysis providing a sufficient condition for the ESP to hold for the \model{} model, as stated in the proposition below.
\begin{prop}
\label{prop:suff_cond}
Let us assume that $ |\phi'|\leq 1$, e.g., $\phi=\tanh$.
If $  \sigma  < 1 $ then the \model{} has the ESP for all inputs.
\end{prop}
\proof\\
Following the same proof of \cite[Proposition 3]{jaeger2001echo}, for an \model{} to have the ESP for all input, it is sufficient to demonstrate that $ \lVert \dfrac{\partial G}{\partial x} (u,x)\rVert <1 $ for all $x, u$.
The following holds
$$
\lVert \dfrac{\partial G}{\partial x} (u,x)  \rVert  \leq 
\beta \lVert  \mathbf{D}_{(u,x)}  \rVert  \lVert \rho\mathbf{W}_{r}  \rVert  + (1-\beta) \lVert \mathbf{O}  \rVert \leq
\beta  \sigma  + (1-\beta) = 
1 - \beta(1- \sigma ).
$$
The first inequality is the triangle inequality, the second holds since $ \lVert  \mathbf{D}_{(u,x)}  \rVert \leq 1$ for all activation functions such that $ |\phi'|\leq 1$, and because of the isometric property of orthogonal matrices which implies $ \lVert \mathbf{O}  \rVert = 1 $. 
Now the thesis holds by hypothesis since $
    \sigma < 1  \Longrightarrow 1 - \beta(1- \sigma) < 1 $.
\qed\\

\begin{remark}
Proposition \ref{prop:suff_cond} informs us that, regardless of the particular value of $\beta \in (0,1] $, the \model{} model owns the ESP under the same contractivity condition of a standard ESN.
\end{remark}

In an \model{} with a relatively low value of $\beta$, the dependence from the hyperparameters $\rho$ and $ \omega $ is attenuated. In fact, the forward dynamics of such an \model {} evolve by design close to the edge of chaos. 
Since the Jacobian of the \model {} map in eq.~\eqref{eq:jacob_ESESN} is itself a convex combination, we can weight more the orthogonal part of the \model{} approaching the hyperparameter $\beta $ to $0$, which by a continuity argument tune the eigenvalues of the Jacobian to stay closer to the unitary circle, pretty much regardless of the spectral radius of the reservoir and the scaling of the input matrix (provided that the spectral radius is bounded). 
%Note that pushing $\alpha $ to zero in a standard Leaky ESN would lead the eigenvalues to converge towards $(1,0)$, while in a ESESN, when pushing $\beta $ to zero, the eigenvalues converge on the unitary circle.
We formalise this intuition exploiting the Bauer-Fike theorem \cite{bauer1960norms}, that we report here for ease of comprehension.
\begin{thm}[Bauer-Fike]
\label{thm:fike}
Let $ \mathbf{A} $ be a diagonalisable matrix, and let $ \mathbf{V}  $ be the eigenvector matrix such that $ \mathbf{A} = \mathbf{V} \mathbf{\Lambda} \mathbf{V}^{-1}  $ where $ \mathbf{\Lambda} $ is the diagonal matrix of the eigenvalues of $ \mathbf{A}  $.
Let $ \mathbf{E} $ be an arbitrary matrix of the same dimension of $ \mathbf{A} $. Then, for all $\mu$ eigenvalues of $ \mathbf{A} + \mathbf{E} $, there exists an eigenvalue $\lambda $ of $ \mathbf{A} $ such that 
\begin{equation}
    \label{eq:fike}
    | \lambda - \mu | \leq \lVert \mathbf{V} \rVert  \lVert \mathbf{V}^{-1} \rVert \lVert \mathbf{E} \rVert .
\end{equation}
\end{thm}

The Bauer-Fike theorem allows us to derive the following characterisation of the eigenspectrum of the Jacobian of the \model{} map. 
\begin{thm}
\label{thm:annular}
Let us consider an \model{} model whose state update equation is given by eq.~\eqref{eq:state_update}, and recall the definitions of $\sigma$ and $\gamma$ of eq.~\eqref{eq:sigma_reservoir} and eq.~\eqref{eq:gamma}.
%Let us denote $ \mathbf{D}_{(u,x)}  = \diag(\phi'(\xi(u,x))) $, and $ \gamma = \lVert \mathbf{D}_{(u,x)}  \rVert$. 
Then, the eigenspectrum of the Jacobian of the \model{} map is confined in the annular neighbourhood of radius $ \beta \gamma \sigma $ of the circle centered in the origin of radius $ 1 - \beta $. In formulas, for each eigenvalue $ \mu $ of the Jacobian matrix of eq.~\eqref{eq:jacob_ESESN} there exists a $\theta \in [0,2\pi)$ such that 
\begin{equation}
  \label{eq:annular}
  %1-\beta - \beta  \sigma \leq \,\,  | \mu | \,\, \leq 1-\beta +  \beta  \sigma.
  |(1-\beta) e^{i \theta} - \mu | \leq \beta \gamma  \sigma.
\end{equation}
\end{thm}
\proof\\
Define $ \mathbf{A} = (1-\beta) \mathbf{O} $, and $ \mathbf{E} =  \beta \mathbf{D}_{(u,x)}  \rho\mathbf{W}_{r}  $, so that the Jacobian of the \model{} model is $ \dfrac{\partial G}{\partial x} (u,x) = \mathbf{A} + \mathbf{E}  $. The matrix $ \mathbf{O} $ is orthogonal, hence there exists an unitary matrix $\mathbf{V}$ such that $ \mathbf{O} = \mathbf{V} \mathbf{\Lambda} \mathbf{V}^{-1}  $, where $ \mathbf{\Lambda} $ is the diagonal matrix of the eigenvalues of $  \mathbf{O}  $. In particular, each eigenvalue of $  \mathbf{O}  $ is of the kind $e^{i \theta} $, for some argument $ \theta \in [0, 2\pi) $, due to the orthogonality of $  \mathbf{O}  $.  
Therefore, $ \mathbf{A} = (1-\beta) \mathbf{V} \mathbf{\Lambda} \mathbf{V}^{-1}  $, and all eigenvalues of $ \mathbf{A} $ have the form $ (1-\beta) e^{i \theta} $, for some argument $ \theta \in [0, 2\pi) $.
In other words, all the eigenvalues of $ \mathbf{A} $ lie on the complex circle centered in the origin with radius $ (1-\beta) $. 
Now, since $ \mathbf{V} $ is unitary, we have that $ \lVert \mathbf{V} \rVert = \lVert \mathbf{V}^{-1} \rVert = 1  $.
Therefore, eq.~\eqref{eq:fike} tells us that for each $\mu$ eigenvalue of $ \dfrac{\partial G}{\partial x} (u,x) $ there exists a complex number $\lambda= (1-\beta) e^{i \theta} $ such that
\begin{equation}
    \label{eq:eigenspectrum}
    |(1-\beta) e^{i \theta} - \mu | \leq \lVert \beta \mathbf{D}_{(u,x)} \rho \mathbf{W}_{r}  \rVert \leq \beta \gamma \lVert \rho\mathbf{W}_{r}  \rVert  = \beta \gamma \sigma.
\end{equation}
This implies that each eigenvalue of $ \dfrac{\partial G}{\partial x} (u,x) $ must be inside the circle of radius $ 1-\beta +  \beta \gamma\sigma $, and outside the circle of radius $ 1-\beta -  \beta \gamma  \sigma $, both centered in zero, which is the thesis. 
\qed\\

A similar argument leads to the following characterisation of a leaky ESN's eigenspectrum.
\begin{corol}
Let us consider a leaky ESN model whose state update equation is given by eq.~\eqref{eq:leaky_esn}.
Then, the eigenspectrum of the Jacobian of the leaky ESN map is confined in a neighbourhood of radius $ \alpha \gamma\sigma $ of the complex number $ (1-\alpha,0)  $. In formulas, for each eigenvalue $ \mu $ of the Jacobian matrix of a leaky ESN's map it holds 
\begin{equation}
  \label{eq:leakyeigenspectrum}
  |(1-\alpha)  - \mu | \leq \alpha \gamma \sigma.
\end{equation}
\end{corol}
\proof
The proof follows the same steps of the proof of Theorem~\eqref{thm:annular}, replacing $\beta$ with $\alpha$, and $ \mathbf{O}$ with the identity matrix $ \mathbf{I}$. Now, since $ \mathbf{A} = (1-\alpha)\mathbf{I}$ has a unique eigenvalue $1-\alpha$ with multiplicity the dimension of $ \mathbf{A} $, then eq.~\eqref{eq:annular} reads $ |(1-\alpha)  - \mu | \leq \alpha \gamma \sigma $, which is the thesis.
\qed\\

\begin{figure}[ht!]
    \centering
    \includegraphics[keepaspectratio=true,scale=0.1]    {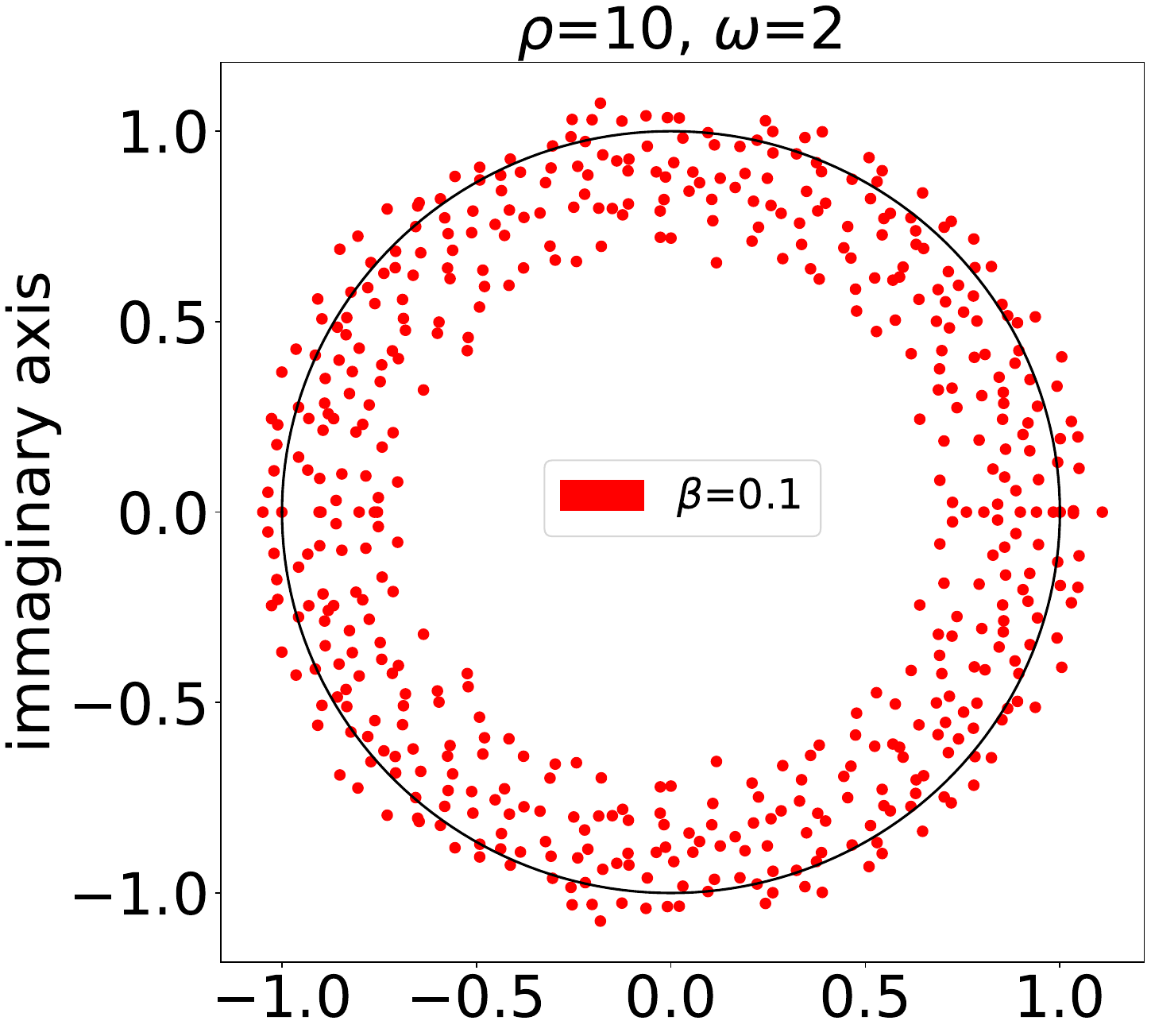}~\includegraphics[keepaspectratio=true,scale=0.1]    {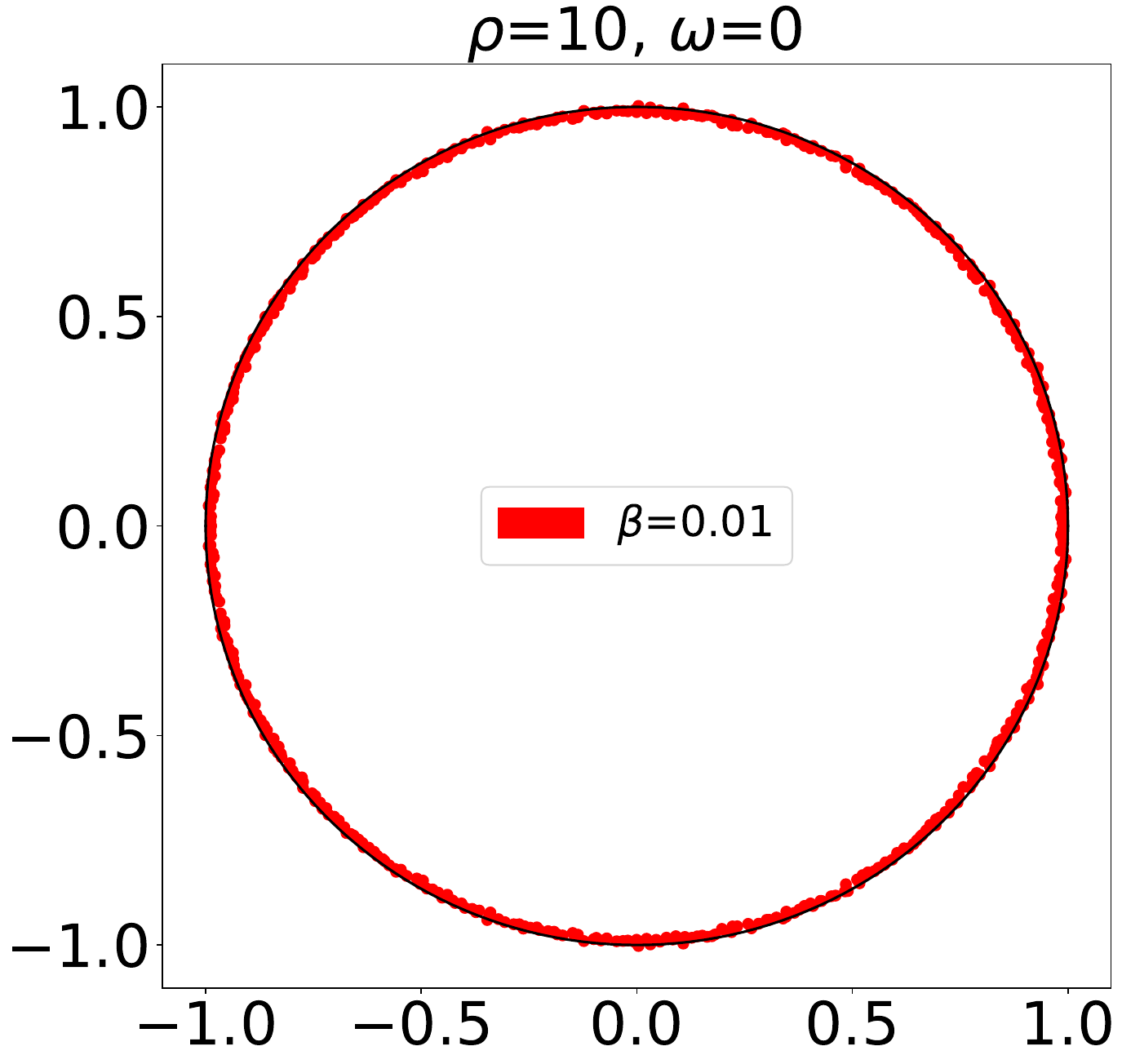}~\includegraphics[keepaspectratio=true,scale=0.1]    {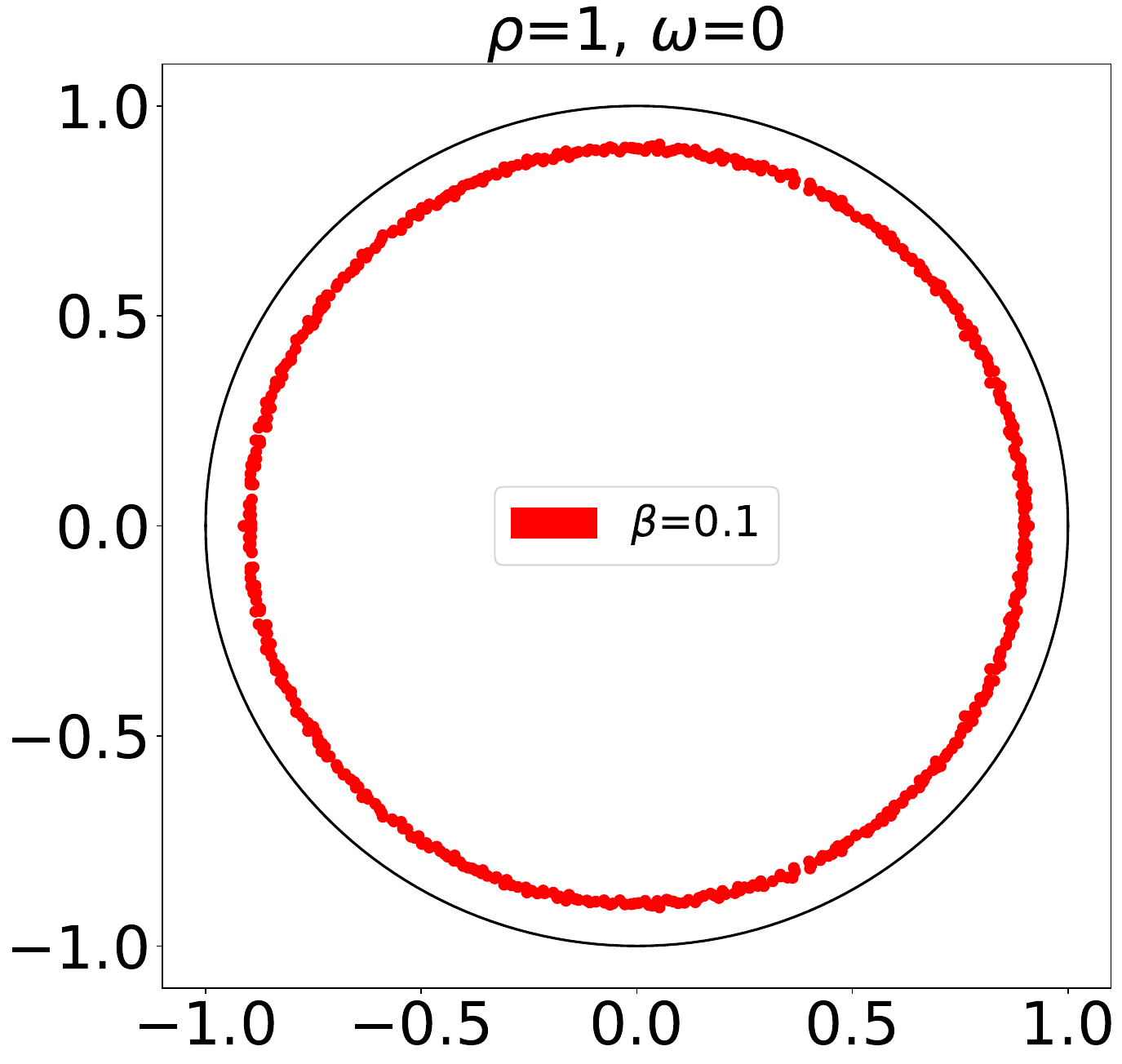}~\includegraphics[keepaspectratio=true,scale=0.1]    {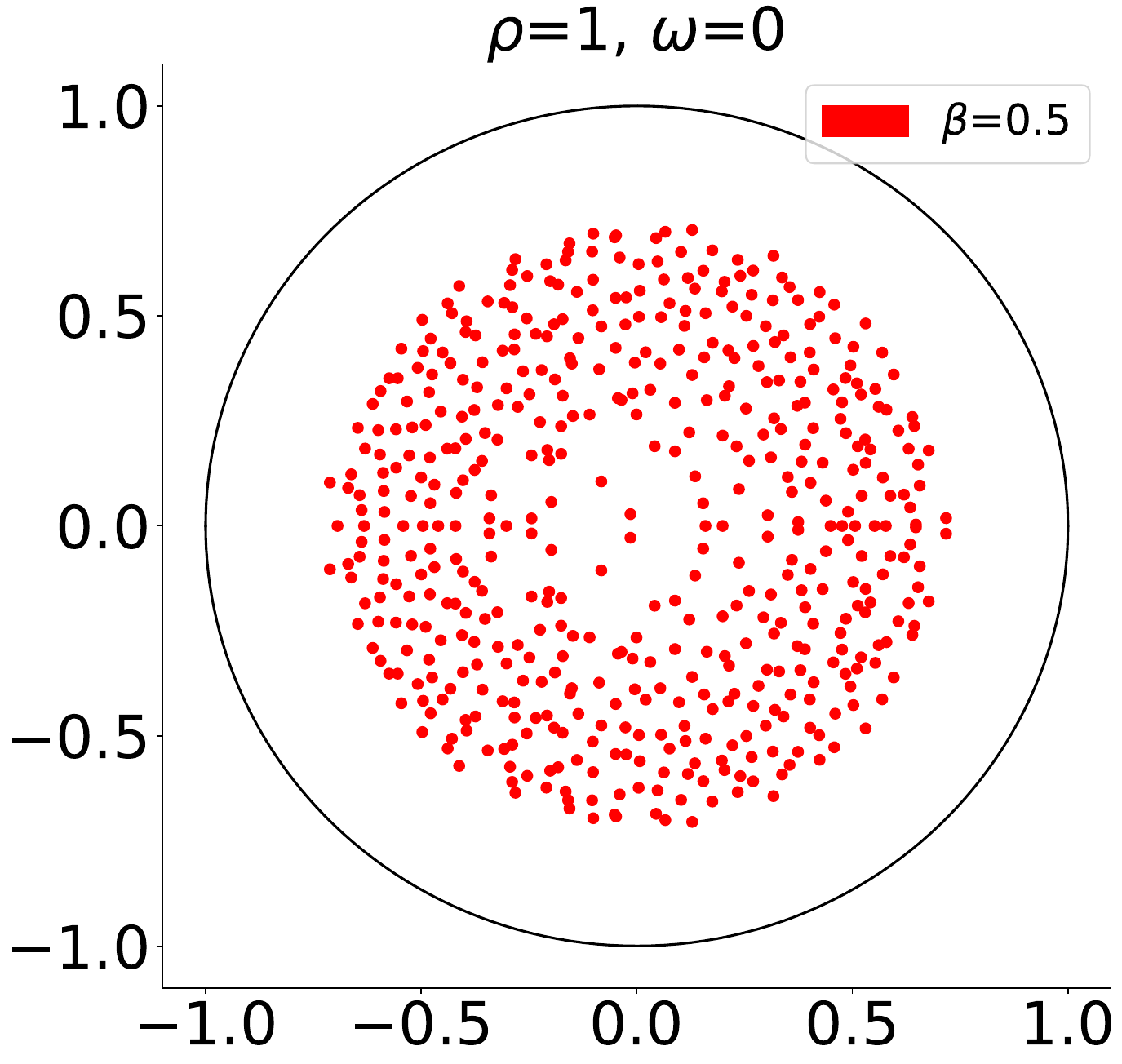}~\includegraphics[keepaspectratio=true,scale=0.1]    {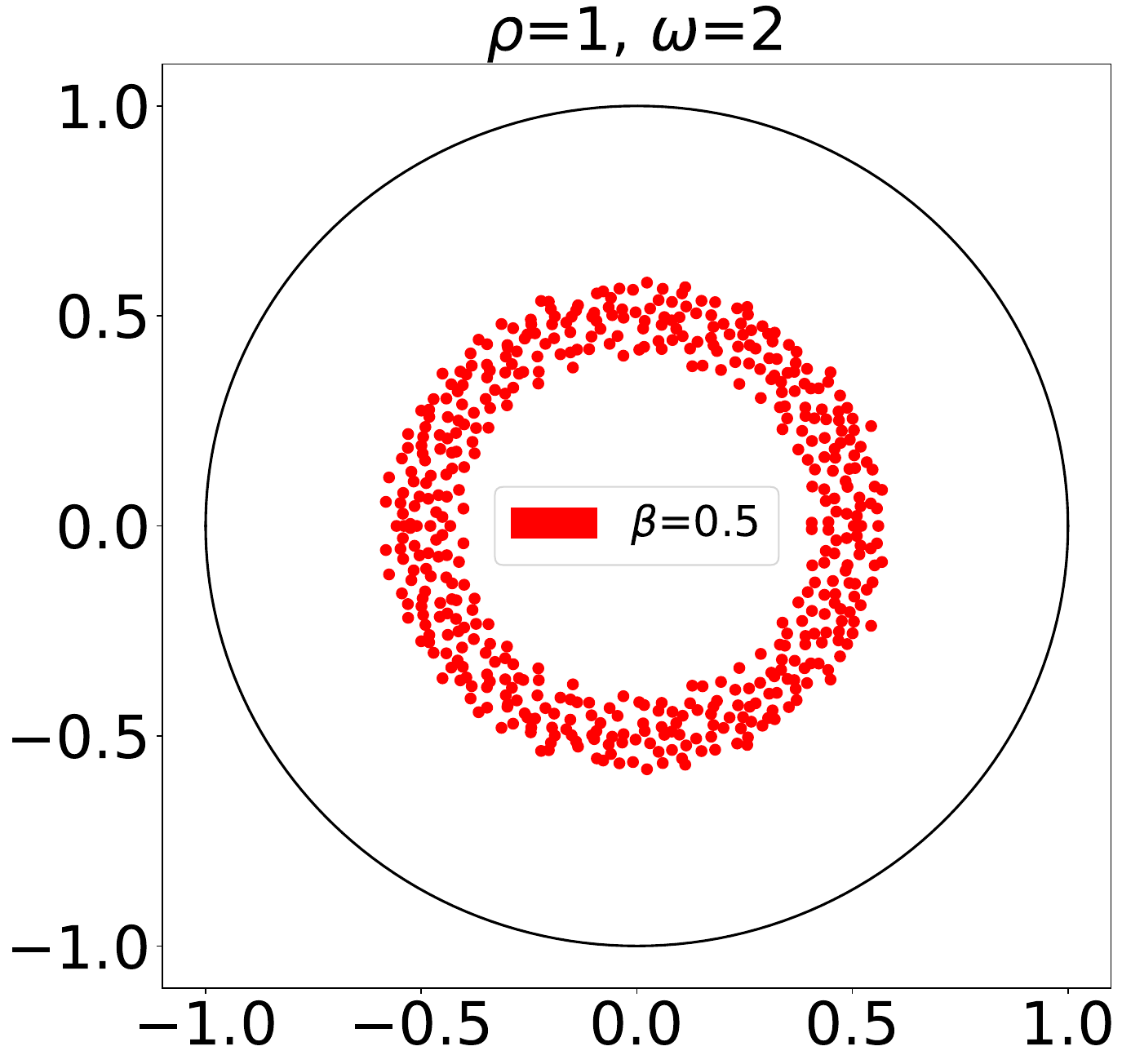}~\includegraphics[keepaspectratio=true,scale=0.1]    {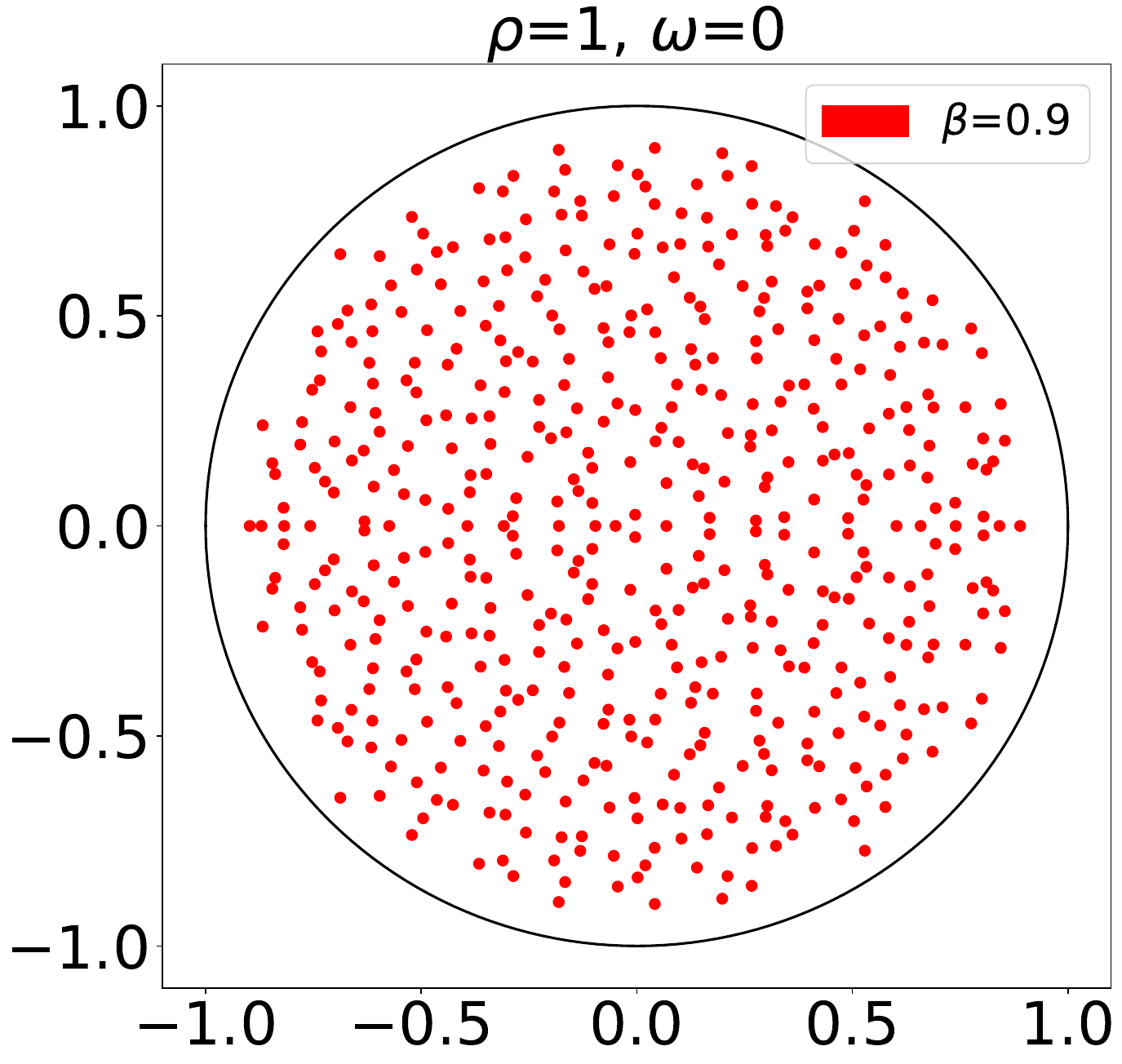}\\
    \includegraphics[keepaspectratio=true,scale=0.1]    {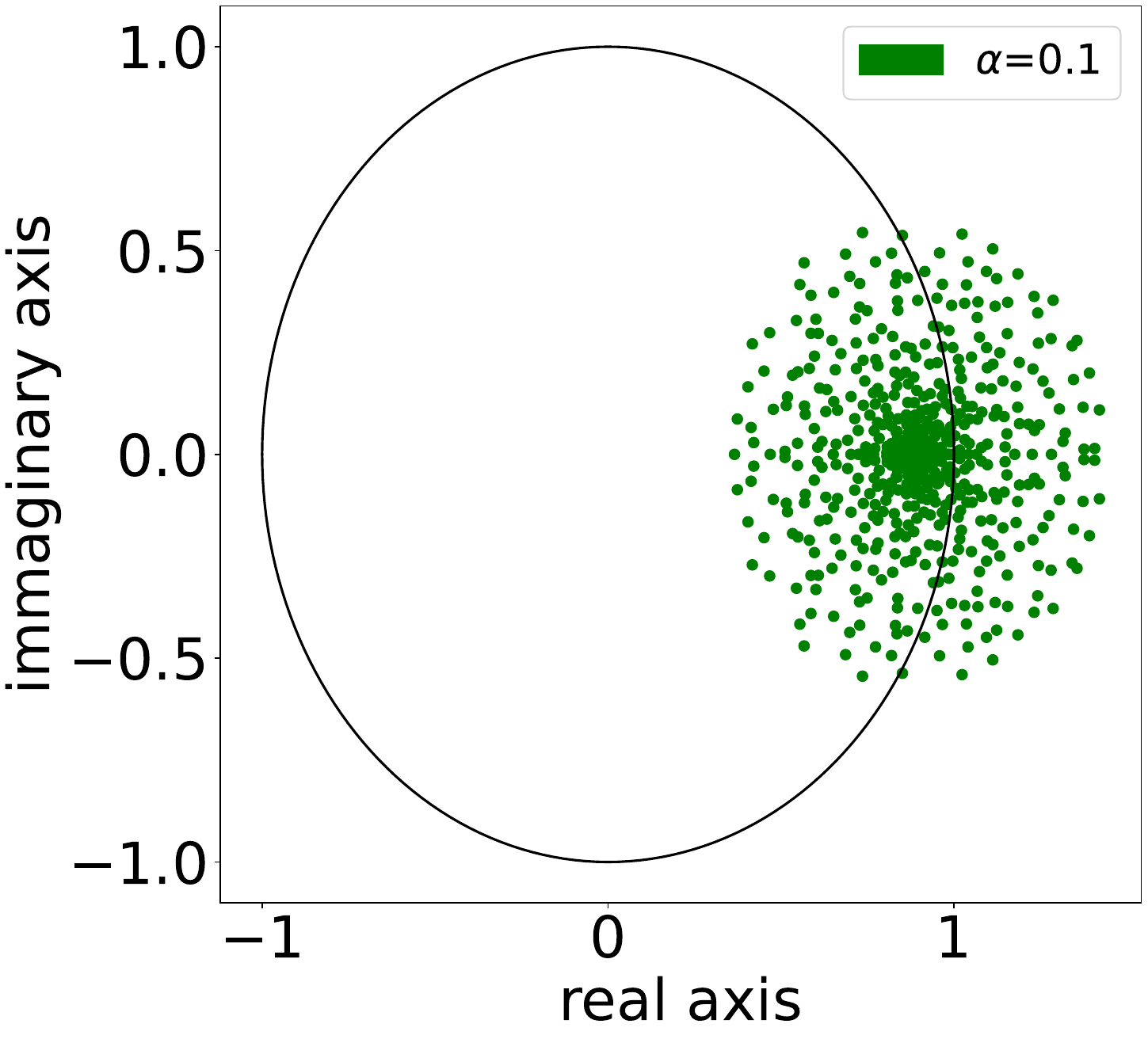}~\includegraphics[keepaspectratio=true,scale=0.1]    {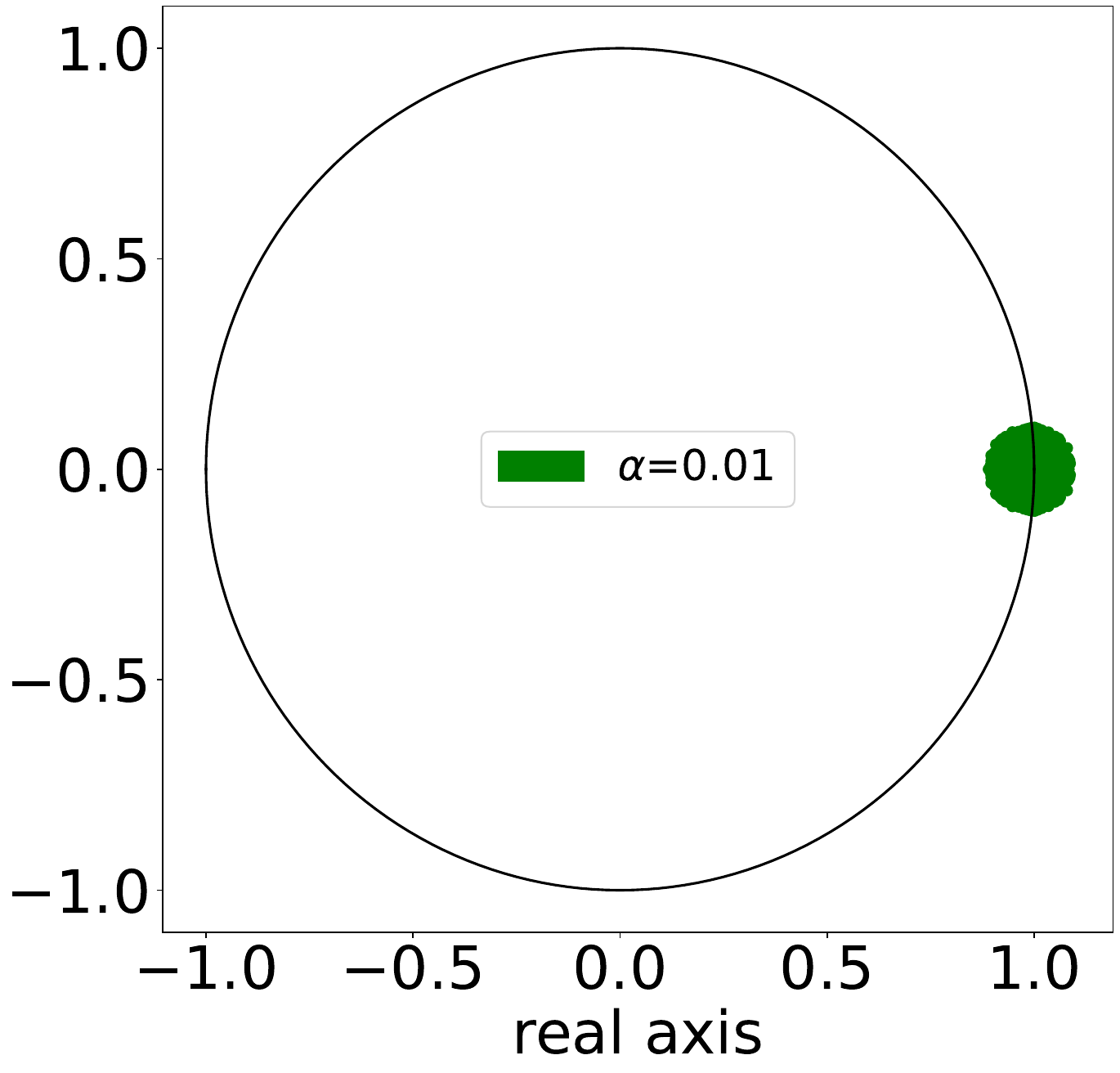}~\includegraphics[keepaspectratio=true,scale=0.1]    {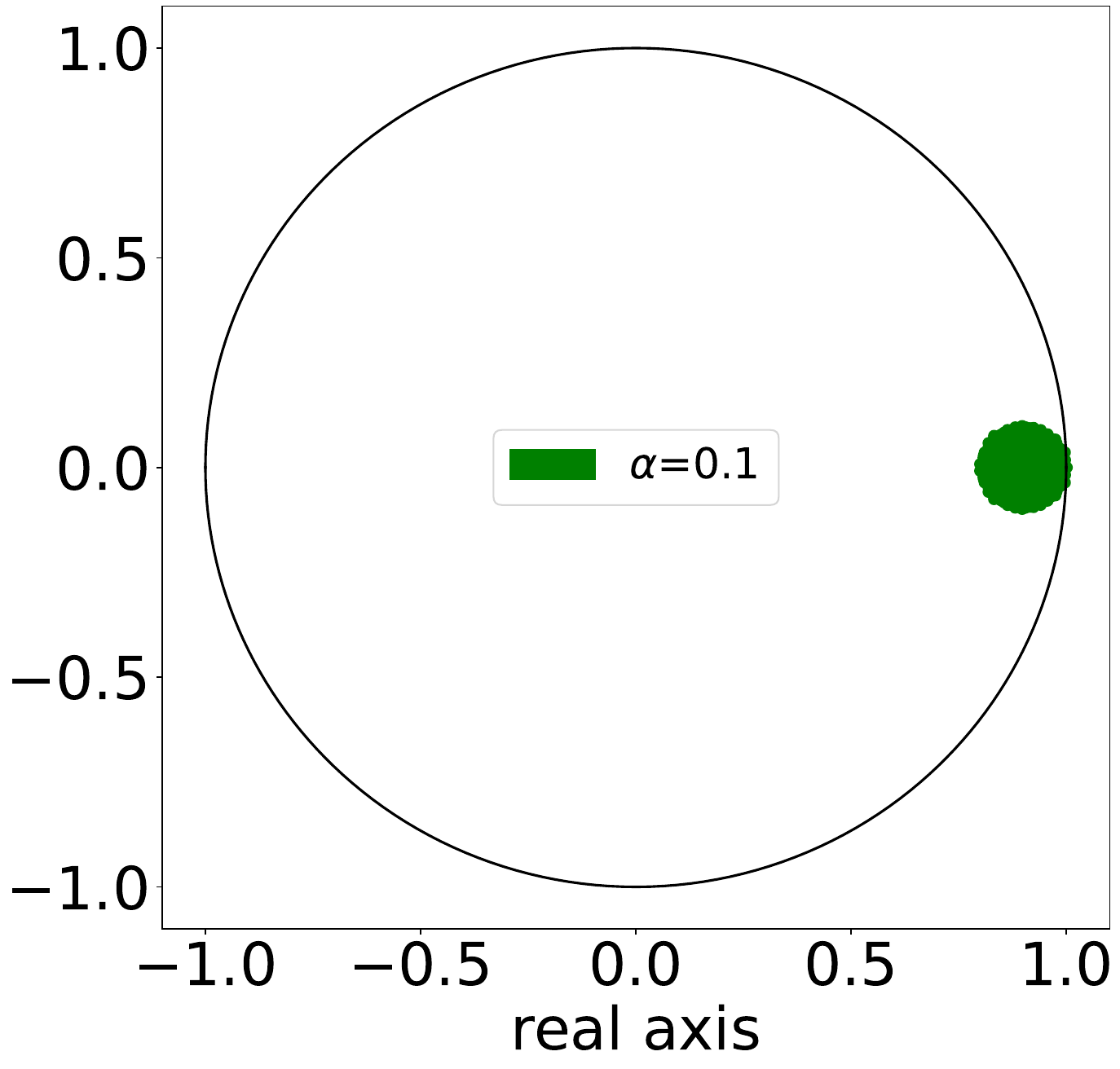}~\includegraphics[keepaspectratio=true,scale=0.1]    {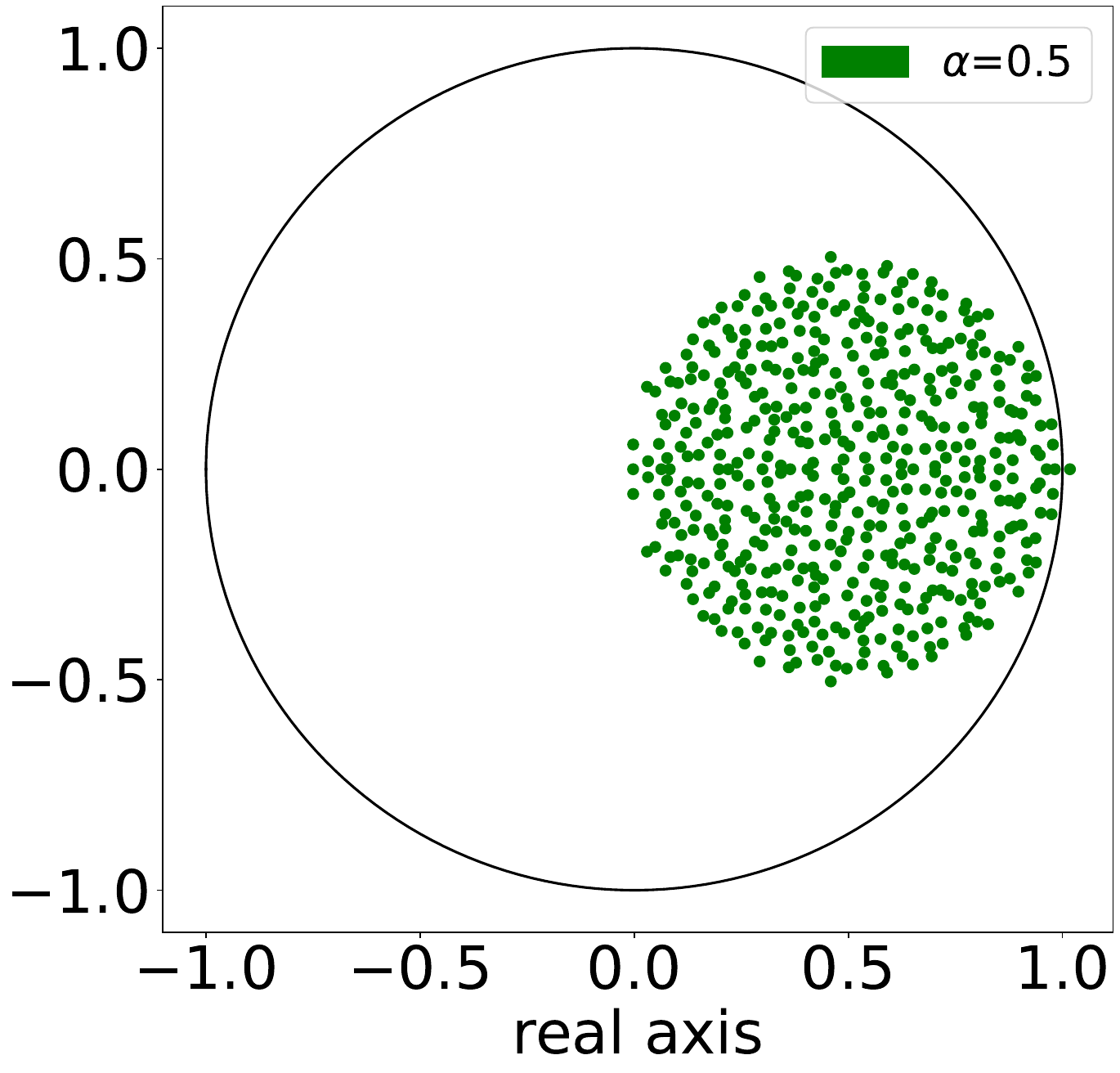}~\includegraphics[keepaspectratio=true,scale=0.1]    {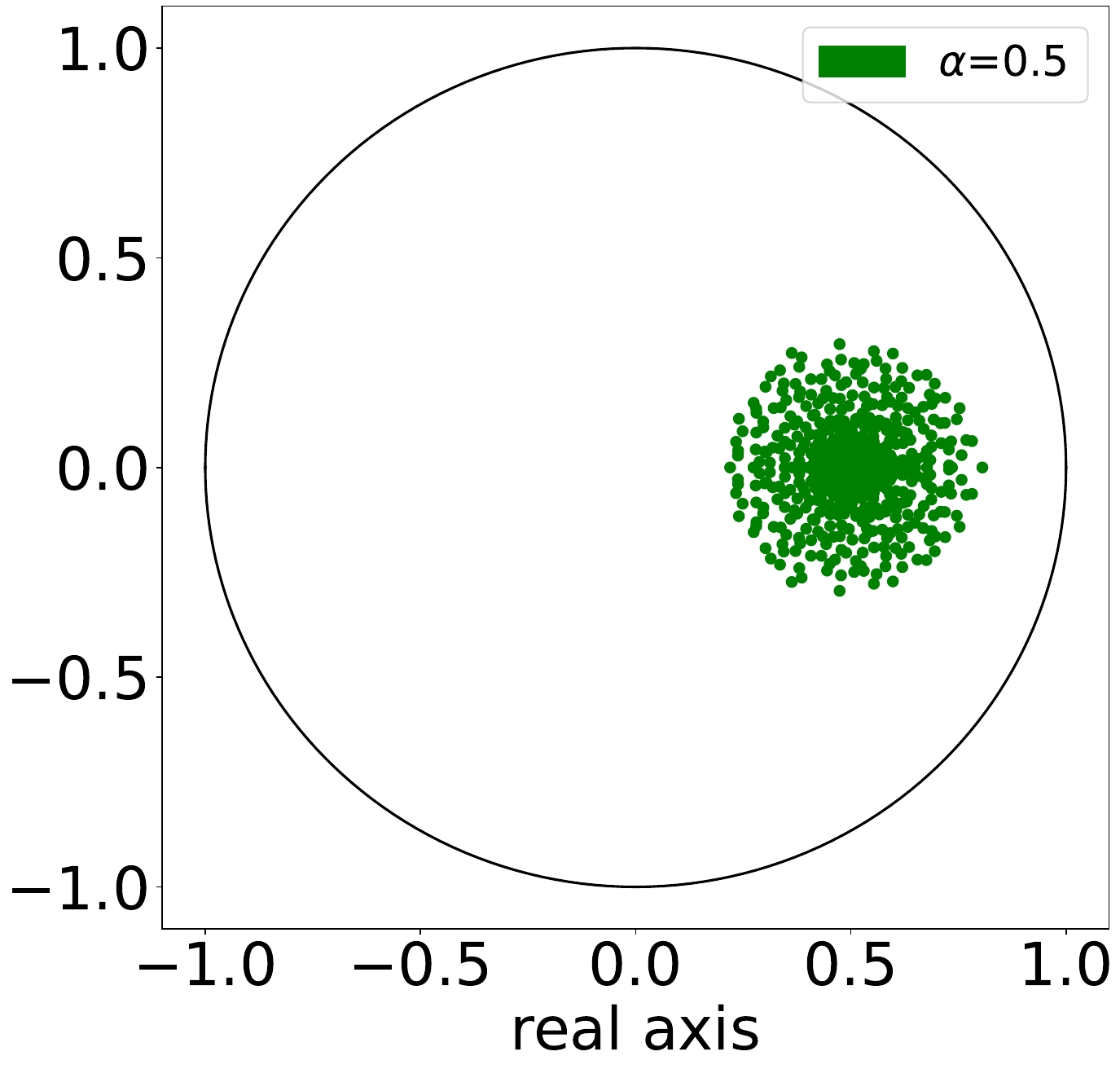}~\includegraphics[keepaspectratio=true,scale=0.1]    {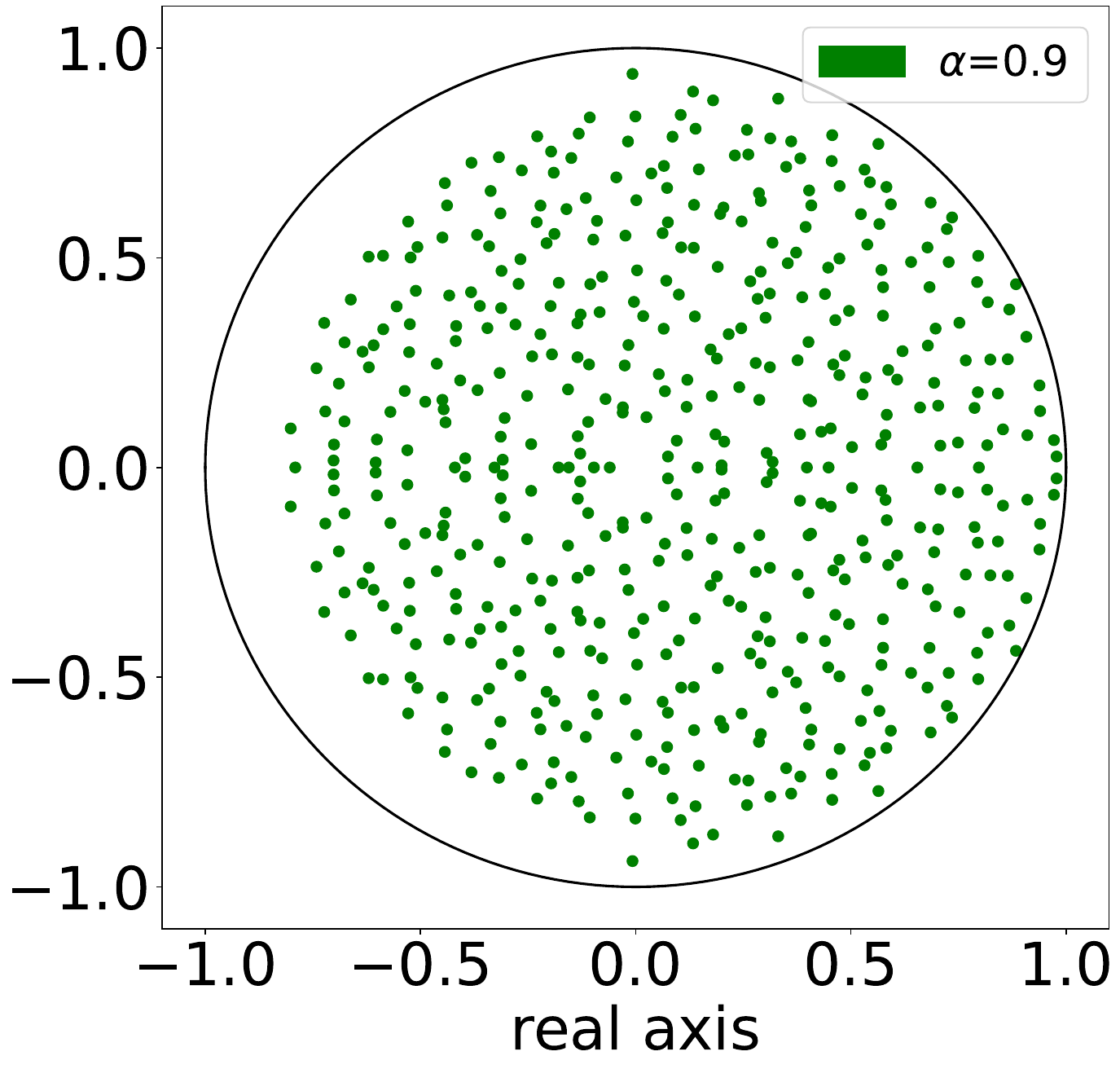}
\caption{Eigenspectrum of the Jacobian of the proposed \model{} model of eq.~\eqref{eq:state_update} (in red) and the leaky ESN model of eq.~\eqref{eq:leaky_esn} (in green) for various combinations of $\rho, \omega, \beta $, and $\alpha$. In black the unitary complex circle. %From left to right values of $[\rho, \omega, \beta \text{ (or $\alpha$)}]$ are $ [10, 2, 0.1], [10, 0, 0.01], [1,0,0.1], [1,0,0.5], [1,2,0.5], [1,0,0.9]  $, where the third value is $\beta$ for the for \model{}, and $\alpha$ for the leaky ESN. 
The input driving the reservoirs is set to the constant $1$, which is then scaled by the value of the hyperparmameter $\omega$.}
\label{fig:eigenspectrum}
\end{figure}

\begin{remark}
\label{rem:different_spectrum}
 The key feature that differentiates an \model{} model from a leaky ESN model is that the latter has an eigenspectrum that shrinks towards $1$ (for small values of $\alpha$), while the former has an eigenspectrum that tends to dispose along the unitary circle (for small values of $\beta$). %We believe that 
 This spread of the eigenspectrum can add richness and diversity to the resulting reservoir dynamics,
 while the “collapse" of the eigenspectrum towards $1$ might harm the expressiveness of the recurrent neural dynamics in some tasks like retrieving memory.
 %, while spreading the eigenspectrum around the unitary circle can add richness and diversity to the resulting reservoir dynamics. 
 In Fig. \ref{fig:eigenspectrum} the eigenspectrum of the Jacobian of the \model{} model and leaky ESN model are plotted for various combinations of the hyperparameters $\rho, \omega, \beta $ and $\alpha$.
\end{remark}

In the context of autonomous dynamical systems, the notion of Maximum Lyapunov Exponent (MLE) is widely adopted to detect whether a system is sensitive to initial conditions \cite{vulpiani2009chaos}. The idea is to consider two infinitesimally close initial conditions and measure the average (over time) maximum expansion rate of the distance between those two initial conditions.
Although there exists a spectrum of Lyapunov Exponents, exactly one for each dimension of the system, the maximum among them is the most important. If the MLE is less than zero, it means that any perturbation of an initial condition gets damped on average over time. On the contrary if the MLE is greater than zero, then there exists at least one direction in tangent space along which the perturbation gets magnified on average over time; this characteristic expansion behaviour of the linearised system is often one of the basic ingredients for the definition of chaotic dynamics \cite{ott2002chaos, strogatz2018nonlinear}.
Therefore, in the literature the edge of chaos is often defined as the locus of parameters where the MLE is exactly zero \cite{boedecker2012information}.
%One way to measure sensitivity to initial conditions in autonomous dynamical systems is to measure the Maximum Lyapunov Exponent (MLE). In the context of autonomous dynamical systems, a widely adopted definition of the Edge of Chaos is given by the condition MLE$=0$. 
However, when we allow an external input to drive the dynamics, as usual in RNNs, the definition of the Lyapunov exponents becomes input-dependent.
In the literature \cite{verstraeten2007experimental,livi2017determination}, one way to assess the degree of regularity of the dynamics of an input-driven system is to compute its Maximum Local Lyapunov Exponent (MLLE) on a given input-driven trajectory, whose definition is provided below.
%Although a clear and widely accepted definition of edge of chaos for nonautonomous dynamical systems does not exist, in the literature \cite{verstraeten2007experimental,livi2017determination}, one way to assess the degree of regularity of the ESN dynamics is to compute its Maximum Local Lyapunov Exponent (MLLE) on a given input-driven trajectory.

\begin{defn}
\label{def:MLLE}
Let us consider an input-driven system of equation $ x[t] = G(u[t], x[t-1] ) $, e.g. the one defined by the \model{}'s map whose state update equation is given by eq.~\eqref{eq:state_update}.
%Let us denote $ \mathbf{D}_{(u,x)}  = \diag(\phi'(\xi(u,x))) $, and $ \gamma = \lVert \mathbf{D}_{(u,x)}  \rVert$. 
Let be given an initial internal state condition $ x[0] $ at time $t=0$, and a sequence of $T$ inputs $ u[1], \ldots, u[T]$. Then the input-driven trajectory $ \{ ( u[t+1], x[t] ) \}_{t=0, \ldots, T-1} $ is well defined via eq.~\eqref{eq:state_update}. Then, the MLLE of the \model{} on such input-driven trajectory is defined as follows
\begin{equation}
    \label{eq:lyapunov}
    \Lambda := \max_{n=1,\ldots, N_r } \dfrac{1}{T} \sum_{t=0}^{T-1} \log( r_n[t] )
\end{equation}
where $N_r$ is the dimension of the reservoir matrix $ \mathbf{W}_{r} $, and $ r_n[t] $ is the square root of the modulus of the $n$th eigenvalue of the symmetric real matrix $\mathbf{J}[t]\mathbf{J}[t]^T$, with $ \mathbf{J}[t] $ defined as in eq.~\eqref{eq:inpdriven_jacob}.
\end{defn}

Roughly speaking, eq.~\eqref{eq:lyapunov} gives us an estimation of the maximum expansion rate, averaged over the window of time $[0,T]$, locally to a given input-driven trajectory. 
On the same vein of autonomous dynamical systems, a $\Lambda<0$ denotes local contractive stable dynamics, $\Lambda>0$ is a blueprint of chaotic dynamics since it implies local exponential divergence of trajectories, while $\Lambda=0$ characterises the edge of chaos.\\
Below we provide an estimation of the MLLE for the \model{} model.
\begin{thm}
\label{prop:edge_chaos}
Let us consider an \model{} model whose state update equation is given by eq.~\eqref{eq:state_update}.
Then, for all time lenghts $T$, initial internal state $x[0]$, and inputs $ u[1], \ldots, u[T]$, the MLLE defined by eq.~\eqref{eq:lyapunov} is bounded as follows
\begin{equation}
\label{eq:lambda_bounds}
\log\bigl( 1 - \beta(\gamma\sigma+1) \bigl) \leq \Lambda \leq \log\bigl(1+ \beta(\gamma\sigma-1) \bigl),
\end{equation}
In particular, in the first order approximation of small values of $\beta$ it holds that
\begin{equation}
    \label{eq:stability}
    \Lambda \approx -\beta .
\end{equation}
\end{thm}
\proof
Thanks to Theorem \ref{thm:annular} we know that the modulus of each eigenvalue $\mu$ of the Jacobian $ \mathbf{J}[t]$ is bounded as  $  1-\beta - \beta \gamma \sigma \leq \,\,  |\mu| \,\, \leq 1-\beta +  \beta \gamma \sigma $, regardless of $t$. Each eigenvalue $ \nu $ of $ \mathbf{J}[t] \mathbf{J}[t]^T$ is bounded as $  (1-\beta - \beta \gamma \sigma)^2 \leq \,\,  |\nu| \,\, \leq (1-\beta +  \beta \gamma \sigma )^2 $, regardless of $t$. In particular, the square root $ r_n[t] $ of the modulus of the $n$th eigenvalue of the matrix $\mathbf{J}[t]\mathbf{J}[t]^T$ is bounded as $  1-\beta - \beta \gamma \sigma \leq \,\,  r_n[t] \,\, \leq 1-\beta +  \beta \gamma \sigma  $, regardless of $n \in \{1, \ldots, N_r \}$, and $t$. Therefore, it follows \eqref{eq:lambda_bounds} from Definition~\ref{def:MLLE}. In particular, for small values of $\beta \approx 0 $ we have a tight squeeze that justifies the estimation of $ \Lambda $ as the arithmetic mean of the first order approximation of the left bound, $ \log\bigl( 1 - \beta(\gamma\sigma+1) \bigl) \approx  - \beta(\gamma\sigma+1)  $, and the right bound, $ \log\bigl( 1 + \beta(\gamma\sigma-1) \bigl) \approx  \beta(\gamma\sigma-1)  $, which results in the approximation $ \Lambda \approx  -\beta  $.
\qed\\

\begin{remark}
Theorem~\ref{prop:edge_chaos} implies that, in an \model{}, we can tune the recurrent neural dynamics towards the edge of chaos via tuning the proximity hyperparameter $\beta$, regardless of the input. More precisely, for decreasing values of $\beta$ the bounds of eq.~\eqref{eq:lambda_bounds} become tighter, and in the approximation of small values of $\beta$ the \model{} model gets close to the edge of chaos ($\Lambda \approx 0$) while being on average over time in the stable regime characterised by $ \Lambda < 0 $.
\end{remark}

\section{Experiments} 
\label{sec:experiments}

In this section we present our experimental analysis on the \model{} model, in comparison with well-established approaches from the ESN literature.
%perform numerical simulations to test the \model{} model. 
Specifically, the short-term memory capacity is analysed in Section~\ref{sec:MC}, while in Section~\ref{sec:inubushi} we investigate the trade-off between nonlinearity and memory. Finally, in Section \ref{sec:mso} we focus on the autoregressive time series modeling, using the multiple superimposed oscillators case as a reference task.
%the run experiments on the multiple superimposed oscillators task. 

\subsection{Memory Capacity}
\label{sec:MC}

The Memory Capacity (MC) task was introduced by Jaeger in \cite{jaeger2002short} to measure the short-term memory ability of an ESN to retrieve from the pool of internal neuronal activations the past input history. 
The task later became very popular for analysing the computational properties of RC-based models \cite{rodan2010minimum}.
%For instance, in \cite{rodan2010minimum} the authors explored simple architectures that are able to maintain high MC scores \cite{rodan2010minimum}; there they propose the Simple Cycle Reservoir architecture.
A systematic analysis of MC varying various hyperparameters of ESNs can be found in \cite{farkavs2016computational}, in which the authors also propose gradient descent based orthogonalization procedures to increase the MC. In the following, we set an MC experiment similarly to \cite{gallicchio2017deep}. 

%Specifically, 
We consider reservoirs of $N_r = 100$ neurons, with a linear readout trained by ridge regression. The input $u[t]$ is an i.i.d. signal uniform in $[-0.8,0.8]$ of discrete-time length 
$T = 6000$.
%$t=1, \ldots, 6000$. 
The first 5000 time steps are exploited for training (excluding the very first 100 time steps to warm up the reservoir), and the last 1000 time steps are left for test. The task is to reproduce in output a signal $z_k[t]$ that is a delayed version of $k$ time steps of the input signal, i.e., to have $z_k[t] $ as close as possible to $ u[t-k]$. 
The MC score is defined as follows:
\begin{equation}
\label{eq:MC}
MC = \sum_{k=1}^{\infty} MC_k ,
\end{equation}
where $ MC_k $ is the squared correlation coefficient between the output $z_k[t]$ and the target $u[t-k]$, defined as follows:
\begin{equation}
    \label{eq:correlcoeff}
    MC_k  =   
    \dfrac{ \Bigl( <  \bigl(z_k[t]-<z_k[t]>_t \bigl) \bigl( u[t-k]- <u[t-k]>_t \bigl) >_t \Bigl)^2}{  < \Bigl( z_k[t] - <z_k[t]>_t \Bigl)^2 >_t \,\, <  \Bigl( u[t-k] - <u[t-k]>_t \Bigl)^2  >_t  }.
\end{equation} 
Angular bracket in eq.~\eqref{eq:correlcoeff} denotes average over time, and are
%All the averages over time in eq.~\eqref{eq:correlcoeff} are 
calculated with regard to the test session, i.e. for $t=5001, \ldots, 6000$.
Moreover, the calculation of the sum in eq.~\eqref{eq:MC} has been truncated to $k=200$, i.e. twice the reservoir size. This choice of truncating the sum in eq.~\eqref{eq:MC}, widely used in the RC community, makes sense considering that the maximum MC achievable by an $N_r$-dimensional reservoir is $N_r$; for a proof of this fact see \cite[Proposition 2]{jaeger2002short}.
%that the maximum MC obtainable with a reservoir of 100 neurons is 100 \cite{jaeger2002short}.

For the calculations, both ESNs and \model{} have been set with 
%$0\% $ reservoir sparsity (i.e., fully connected), and set with 
a spectral radius of $0.9$, and an input scaling of $0.1$. This setting has been tested as good for ESNs in previous works \cite{gallicchio2017deep, schrauwen2008improving}.
Keeping fixed those hyperparameters, the leak rate $\alpha$ for the ESN model, and the proximity hyperparameter $\beta$ for the \model{} model, have been varied in $(10^{-3}, 1)$.
Precisely, we used the same grid of 50 values for $\alpha$ and $\beta$, generated via the formula $ a10^{-s} $, with $a$ random uniform in $(0.1,1)$, and $s$ random uniform in $\{ 0, 1, 2\}$. This methodology, while covering a large range of values for the leak rate of ESNs, also ensures to properly explore values close to zero where the \model{} model presents an interesting behaviour.
%In fact, a theoretical result \cite[Theorem 3.1]{ceni2022random}, revealed that values in a neighbourhood of $\beta = \dfrac{1}{L}$, where $L$ is the length of a given time series, render an optimal forward and backward propagation of signals through the recurrent neural network. Now, considered that we would like to retrieve information up to 100 time steps in the past, then we expect to find an optimal value around $ \beta=10^{-2}$.

%
\begin{figure}[ht!]
    \includegraphics[keepaspectratio=true,scale=0.3]{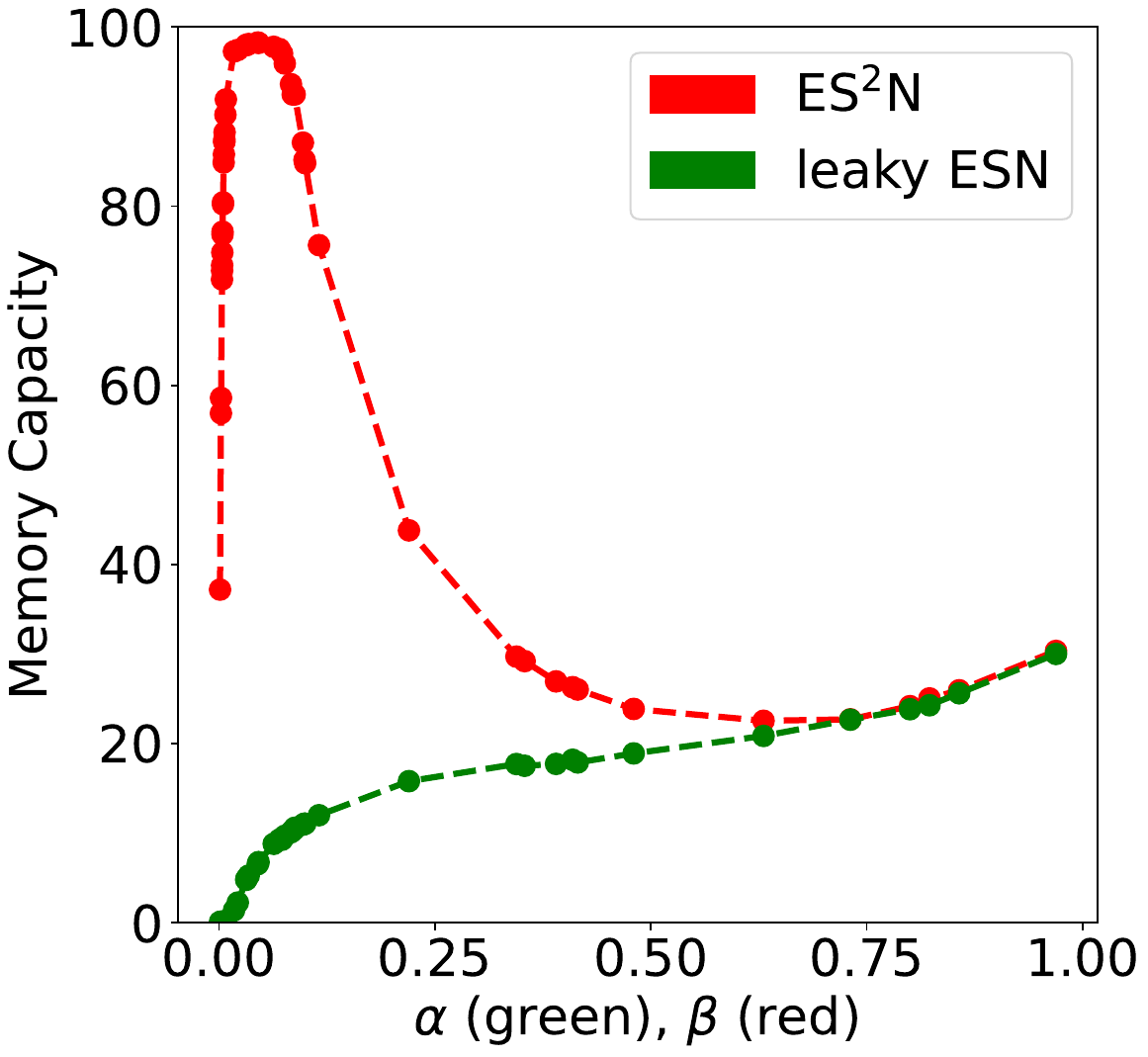}~\hspace{0.2cm}            \includegraphics[keepaspectratio=true,scale=0.375]{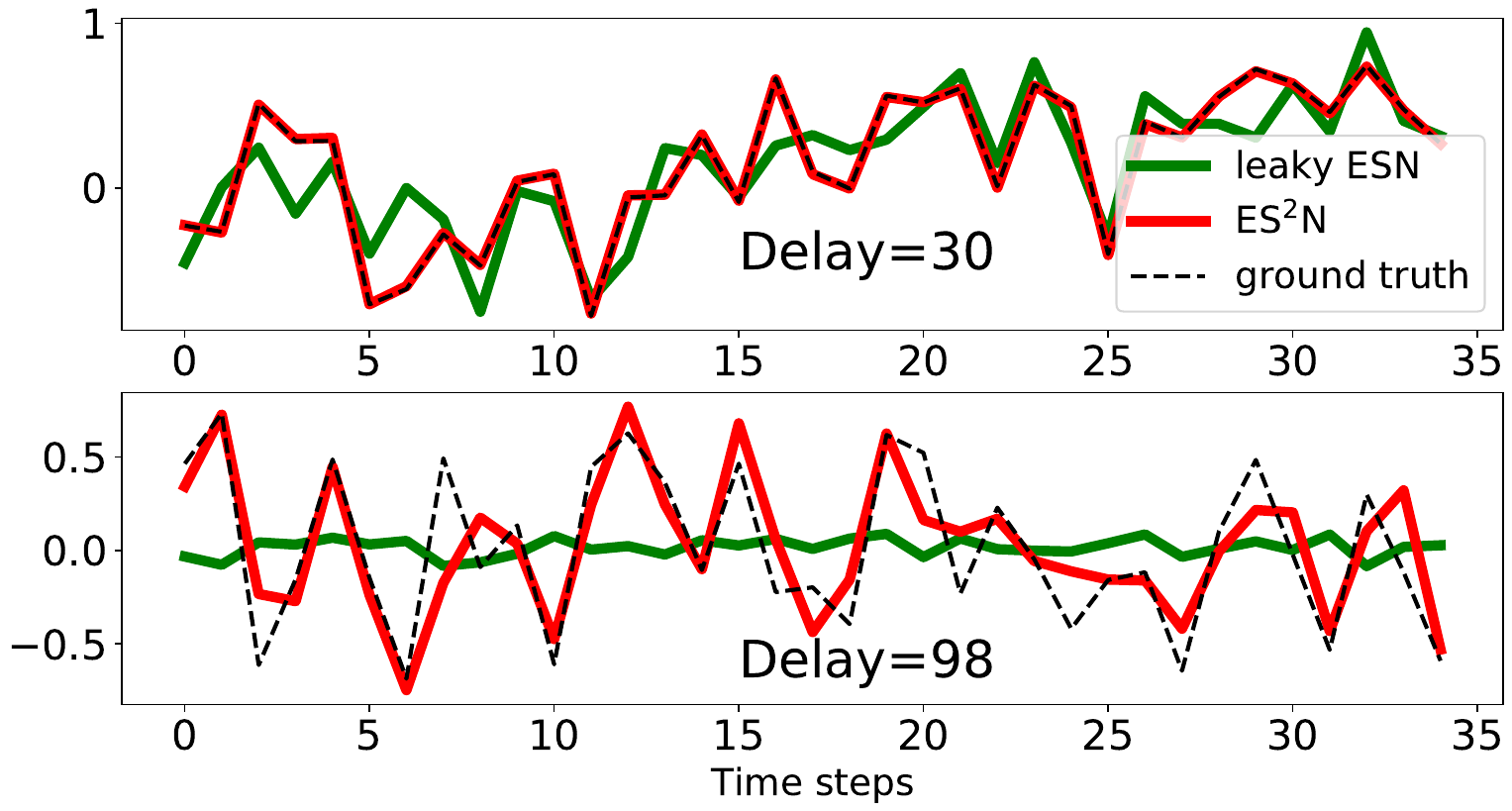}
    \caption{\textbf{Left:} MC values of eq.~\eqref{eq:MC} averaged over 10 trials for various values of $\alpha$ (for the ESN), and $\beta$ (for the \model{}). \textbf{Right:} Output signals of ESN with $\alpha=1$ (green), and \model{} with $\beta=0.05$ (red), over the ground truth signal (dashed), for the two cases of delay $k=30$ and $k=98$.}%
    \label{fig:MC}
\end{figure}
\begin{figure}
    \begin{floatrow}
    
    \capbtabbox{%
        \resizebox{5cm}{!}{
            \begin{tabular}{cc}
            \hline
            \textbf{Model} & \textbf{MC} \\
            \hline\\
            leaky ESN  & $ 30.40 \pm 3.76 $\\
            linearESN  & $ 49.35 \pm 17.13 $\\
            orthoESN & $ 89.42 \pm 1.50 $\\
            \linear{} & $ 99.09 \pm 0.01 $ \\
            \hline 
            \model{}   & $ 98.43 \pm 0.11$\\
            \hline
            \end{tabular}
        }
    }{%
      \caption{Mean and standard deviation of the MC computed over 10 different initialisations of reservoir models of 100 neurons. Leaky ESN is with $\alpha=1$, \model{} is with $\beta=0.05$}%
      \label{tab:MC}
    }
    
    %\hspace{0.2cm}
    
    \ffigbox{%
        \includegraphics[keepaspectratio=true,scale=0.185]
        {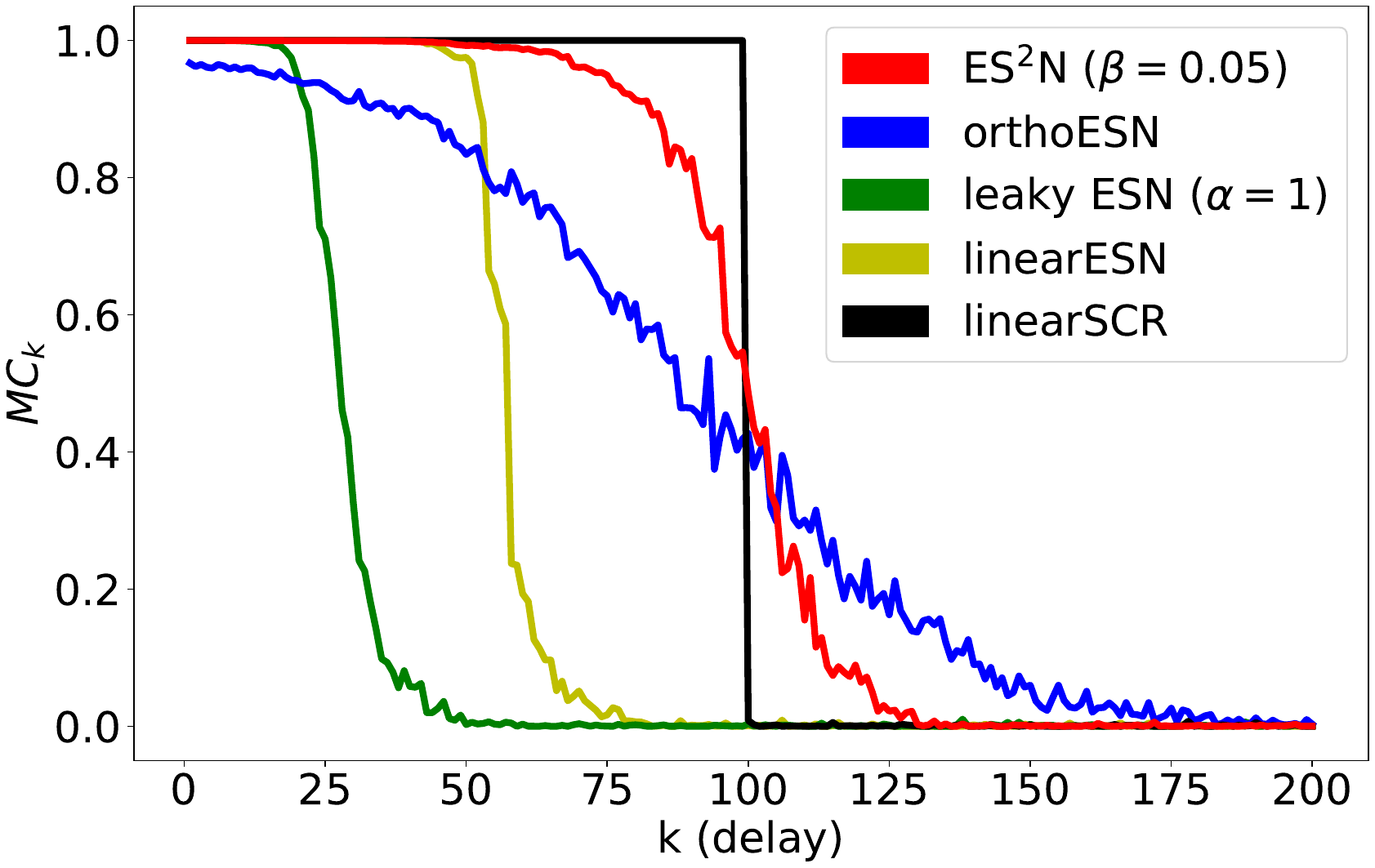}~
        \includegraphics[keepaspectratio=true,scale=0.185]{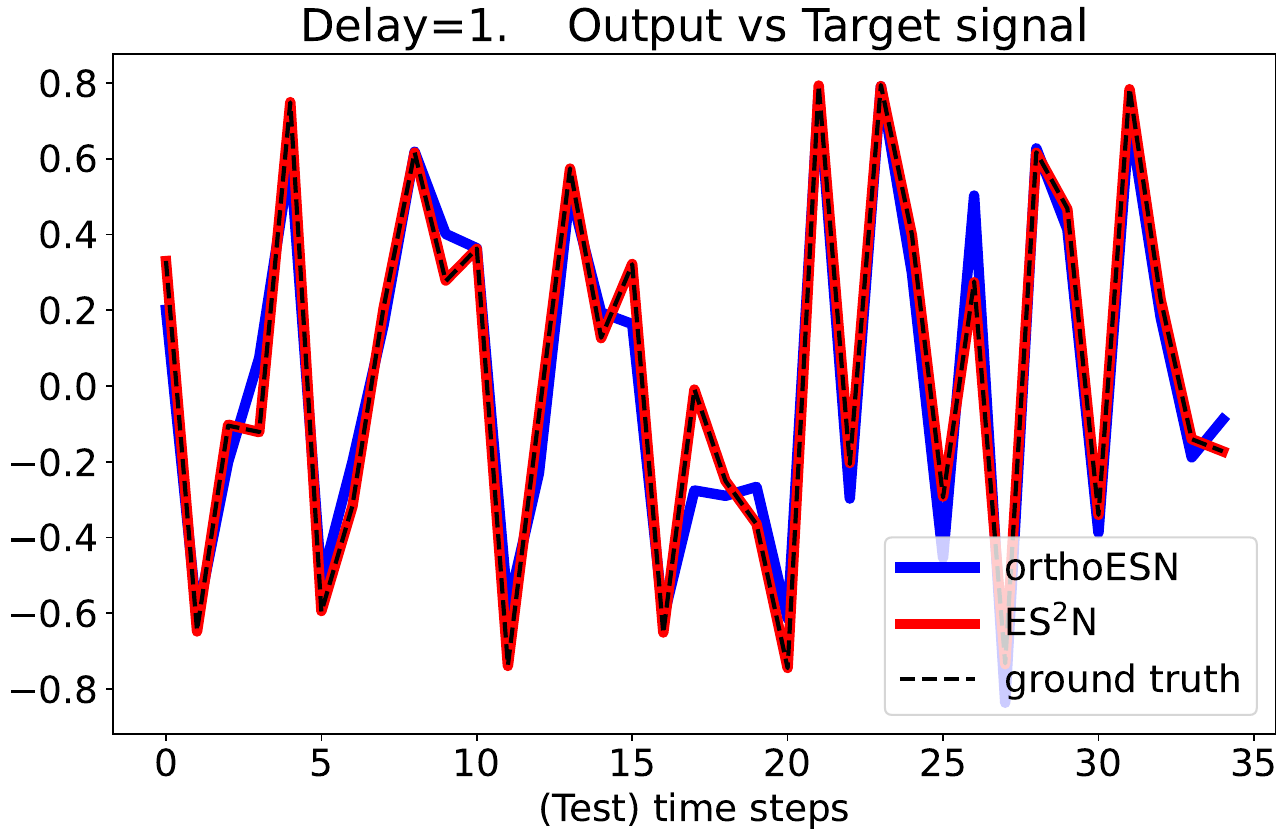}
        }{%
          \caption{\textbf{Left:} $MC_k$ values for delays $k=1,\ldots, 200$, for all the 5 models.\\ \textbf{Right:} orthoESN versus \model{} in the simplest case of $k=1$.}%
          \label{fig:MC_varying_k}
        }

    \end{floatrow}
\end{figure}

Both ESNs and \model{}s have been run for 10 different initialisations for each delay $k$, and the computed MC has been averaged over these trials in order to have statistical significance.
In the left picture of Figure \ref{fig:MC} the computed MC values for both ESN and \model{} are plotted. The MC of \model {}  exhibits a peculiar nonlinear dependence on the proximity hyperparameter $\beta$, peaking around $\beta=0.05$.
On the contrary, the MC of ESN appears monotonic in $\alpha$, reaching the highest value at $\alpha=1$.
As expected, for $\beta$ values approaching to 1, %the \model{}'s MC converges to the ESN's MC. Therefore, it is not surprising to see 
the \model{}'s MC and the ESN's MC curves overlap on the right part of the graph. %Contrariwise, the different shapes of the MC curves for smaller values of $\beta$ and $\alpha$ testify the different functionality of those hyperparameters. 
%For this reason, it does not sound fair to interpret the scalar $\beta$ as a leak term, and that is why in this paper we preferred to use different Greek letters for them. 
In the right 
%pictures 
plots
of Figure \ref{fig:MC}, two examples with delay $k=30$ and $k=98$ are reported for leaky ESN and \model{} (each with its best hyperparameter setting found) highlighting the supremacy of the \model{} model over a standard ESN. Note that recalling the input signal up to 98 time steps in the past is challenging for a reservoir of 100 units.
The computed mean MC values for the optimal $\alpha$ and $\beta$ are reported in Table \ref{tab:MC} along with their empirical standard deviations. 

Additionally, we computed the MC of linear ESN, i.e. the model of eq.~\eqref{eq:leaky_esn} with $\alpha=1$ and the identity function as $\phi$ (called linearESN in Table \ref{tab:MC}); the MC of an ESN with an orthogonal reservoir, i.e. the model of eq.~\eqref{eq:leaky_esn} with $\alpha=1$ and a randomly generated orthogonal matrix $\mathbf{W}_r$ (called orthoESN in Table \ref{tab:MC}); and the MC of a linear ESN with a specific orthogonal structure that realises a circular shift (called \linear{} in Table \ref{tab:MC}), i.e. the model of eq.~\eqref{eq:leaky_esn} with $\alpha=1$ and $\mathbf{W}_r$ with nonzero elements in the lower subdiagonal and the upper-right corner, all filled with $1$.\footnote{The \linear{} model implements the same reservoir topology used in the \emph{Simple Cycle Reservoir} in \cite{rodan2010minimum}, from which the name linearSCR.}  %............ and the MC of a linear ESN with an orthogonal reservoir (linearOrthoESN in Table \ref{tab:MC}), i.e. the model of eq.~\eqref{eq:leaky_esn} with $\alpha=1$, a randomly generated orthogonal matrix $\mathbf{W}_r$, and the identity function as $\phi$. 
All the other hyperparameters of linearESN, orthoESN, and \linear{} have been set identically to those specified previously.
%
%The reservoir size has been proved to be an upper bound for the MC, for ESNs driven by i.i.d. input signals and a linear readout. Moreover, \cite[Proposition 4]{jaeger2002short} gives us a characterisation of linear ESNs with optimal MC. However, it is hard to reach this optimal value, probably due to the effect of numerical error accumulation. This effect is especially evident in linear systems where the errors amplify unboundedly. 
Linear ESNs are known to perform better than nonlinear ESNs in the MC task.
However, as evident from the large standard deviation of linearESN in Table \ref{tab:MC}, linear models occasionally give very poor performance. 
One way to stabilise linear ESN's performance is to employ an orthogonal matrix as reservoir. We used \linear{} as benchmark because theoretical results (see \cite[Proposition 4]{jaeger2002short} and \cite[Theorem 1]{rodan2010minimum}) guarantee for it optimal performance on the MC task.
Remarkably, the \model{} model can get very close to the MC optimal value (of 100, for a reservoir of 100 neurons) with a noticeably narrow standard deviation, almost matching the MC of \linear{}, see Table~\ref{tab:MC}.
%Moreover, a theoretical result \cite[Proposition 4]{jaeger2002short}, for i.i.d. input signals and a linear readout, gives us a necessary and sufficient condition of linear ESNs with optimal MC. However, it is hard to reach this optimal value, probably due to the effect of numerical error accumulation. This effect is especially evident in linear systems where the errors amplify unboundedly. 
%As evident from the large standard deviation of linear ESN in Table \ref{tab:MC}, linear models occasionally give very poor performance. On the contrary, the \model{} model can get very close to the MC optimal value even in the presence of strong noise.

%For the sake of completeness, we measured the MC in the presence of strong noise, following a similar experiment found in \cite{jaeger2002short}. We ran another simulation where we added a source of Gaussian noise $\eta \in N(0, 0.1)$ inside the argument of the activation function of eq.~\eqref{eq:state_update}. Therefore, the \model{}' state-update equation during training was
%\begin{equation}
%    \label{eq:noisy_ESESN}
%    x[t] = \beta  \phi(\mathbf{W}_{r} x[t-1] +  \mathbf{W}_{in}u[t] + \eta[t]) + (1-\beta) \mathbf{O} x[t-1].
%\end{equation}
%The computed \model{}'s MC was $98.40 \pm 0.20$, which demonstrates the resilience of the \model{} model to perturbations. The same experiment on linear ESN caused an higher percentage of fails, resulting in an MC of $44.90 \pm 22.39$.

%Concluding, we want to emphasise how deceptive is the good performance of orthoESN.
Finally, we plot in the left picture of Figure \ref{fig:MC_varying_k} the squared correlation coefficient $MC_k$ for all $k=1, \ldots, 200$, for all of the five considered models: leaky ESN (with $\alpha=1$, the best found), linearESN, orthoESN, \linear{}, and \model{} (with $\beta=0.05$, the best found). These $MC_k$ values are not averaged over more trials. However, apart from linearESN (which sometimes fails), the $MC_k$ values of all the other four models are relatively insensitive to the random initialisation.
As evident from the left plot of Figure \ref{fig:MC_varying_k}, \model{} presents an MC curve (red) particularly close to the optimal one of \linear{} (black), while we note that despite the large MC of the orthoESN model, the orthoESN never excels in reconstructing the delayed version of the input, even for the simplest case of $k=1$.
In the right plot of Figure \ref{fig:MC_varying_k} a comparison between orthoESN and \model{} outputs is plotted, for the case of delay $k=1$. The \model{} (red) is able to perfectly reconstruct the target (black dashed line), while the orthoESN (blue) seems to struggle.
Interestingly, the orthoESN trades its poor reconstruction of the delayed signal with the ability to mildly correlate its output with the target for very large delays, and given the definition of $MC=\sum_k MC_k $, this results in an overall large value of memory capacity.

\iffalse
%##### ESESN with SimpleCycle #####
In Appendix \ref{sec:esesn_simplecycle} further experiments have been run on \model{} using circular shifts of various lengths as orthogonal matrix $ \mathbf{O} $, leading to the conclusion that with \model{} we can precisely design the amount of short-term memory needed in our model.
%##### ESESN with SimpleCycle #####
\fi

%

\subsection{Memory-nonlinearity trade-off}
\label{sec:inubushi}

From previous RC literature, it is well known the existence of a trade-off between the strength of nonlinearity of a dynamical reservoir system and its short-term memory abilities \cite{dambre2012information}. 
%In this section, we investigate how the proposed ESESN model deals with this trade-off.
%In the reservoir computing community, it is a well-known fact that linear ESNs can achieve better performance on memory-demanding tasks.
In \cite{inubushi2017reservoir} the authors propose a task with the aim of measuring this trade-off.
%To overcome this issue, in \cite{inubushi2017reservoir} the authors proposed a reservoir with a tunable proportion of linear and nonlinear units. For the purpose of showing the benefits of having both linear and nonlinear units they train ESNs on 
The task consists of extracting from an i.i.d. uniform input signal $u[t]$ in $[-1,1]$ a target signal of the form $y[t]=\sin( \nu * u[t- \tau] ) $; here $\nu$ quantifies the \emph{nonlinearity strength}, while $\tau$ measures the \emph{memory depth}. 
We use this task to benchmark the memory-nonlinearity trade-off for various combinations of $\tau $ and $\nu$ comparing \model{} against leaky ESN, and \linear{}.
For this experiment, an input signal of length 6000 has been generated, of which the first 5000 for training (excluding the very first 100 steps), and the remaining 1000 for testing. The metric used to evaluate the task is the NRMSE between target $y(t)$ and output $z(t)$ in the test session, defined as follows:
\begin{equation}
    \label{eq:nrmse}
    NRMSE(y,z) = \sqrt{ \dfrac{ < \lVert y[t] -z[t] \rVert^2 >_t }{ < \lVert y[t] - <y[t]>_t \rVert^2 >_t } },
\end{equation}
where $\lVert v \rVert$ represents the Euclidean norm of a vector $v$, and angular brackets means the average over time.
According to this metric, the lower the better.\\

\begin{figure}[ht!]
    \centering
    \includegraphics[keepaspectratio=true,scale=0.21]{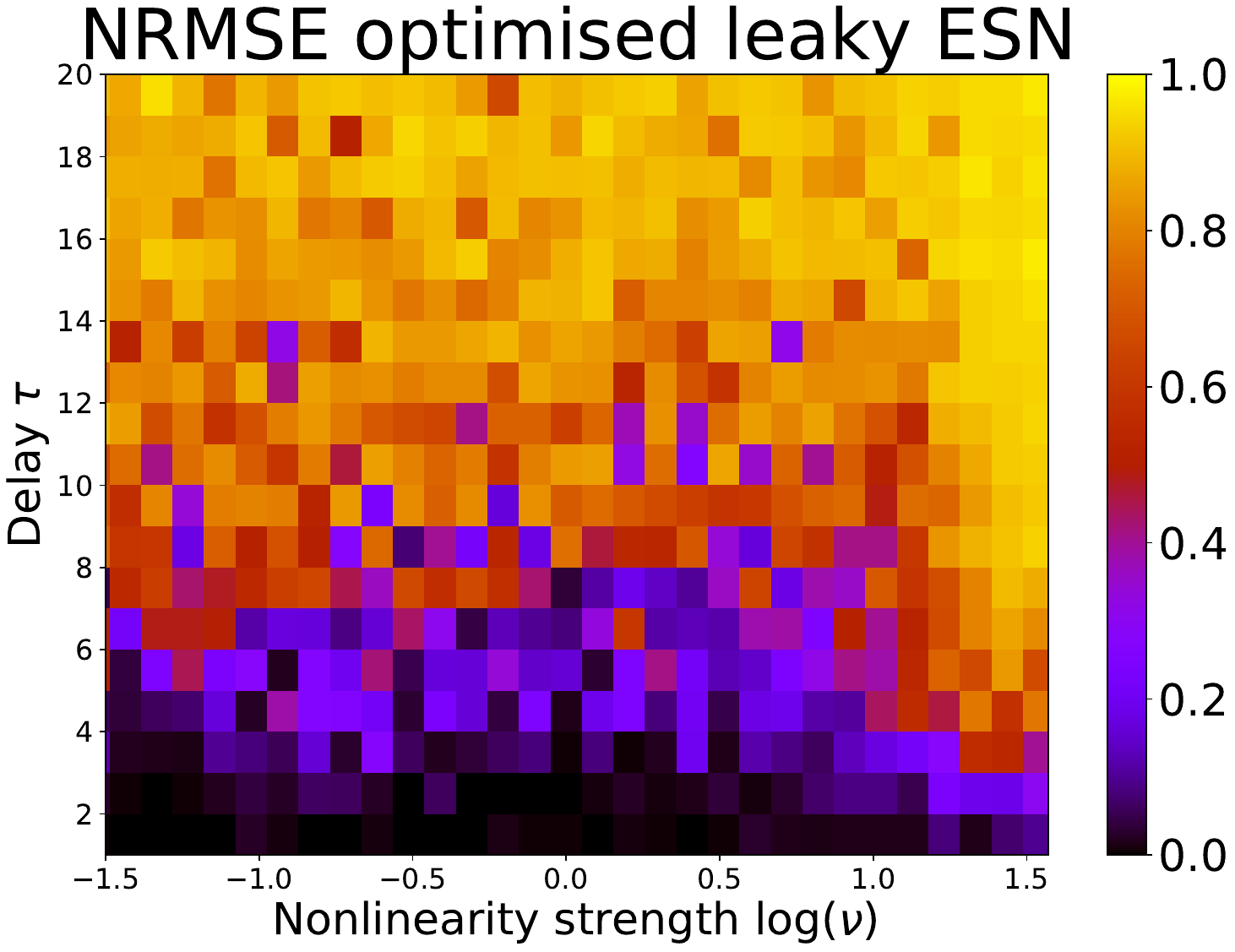}   ~\includegraphics[keepaspectratio=true,scale=0.21]{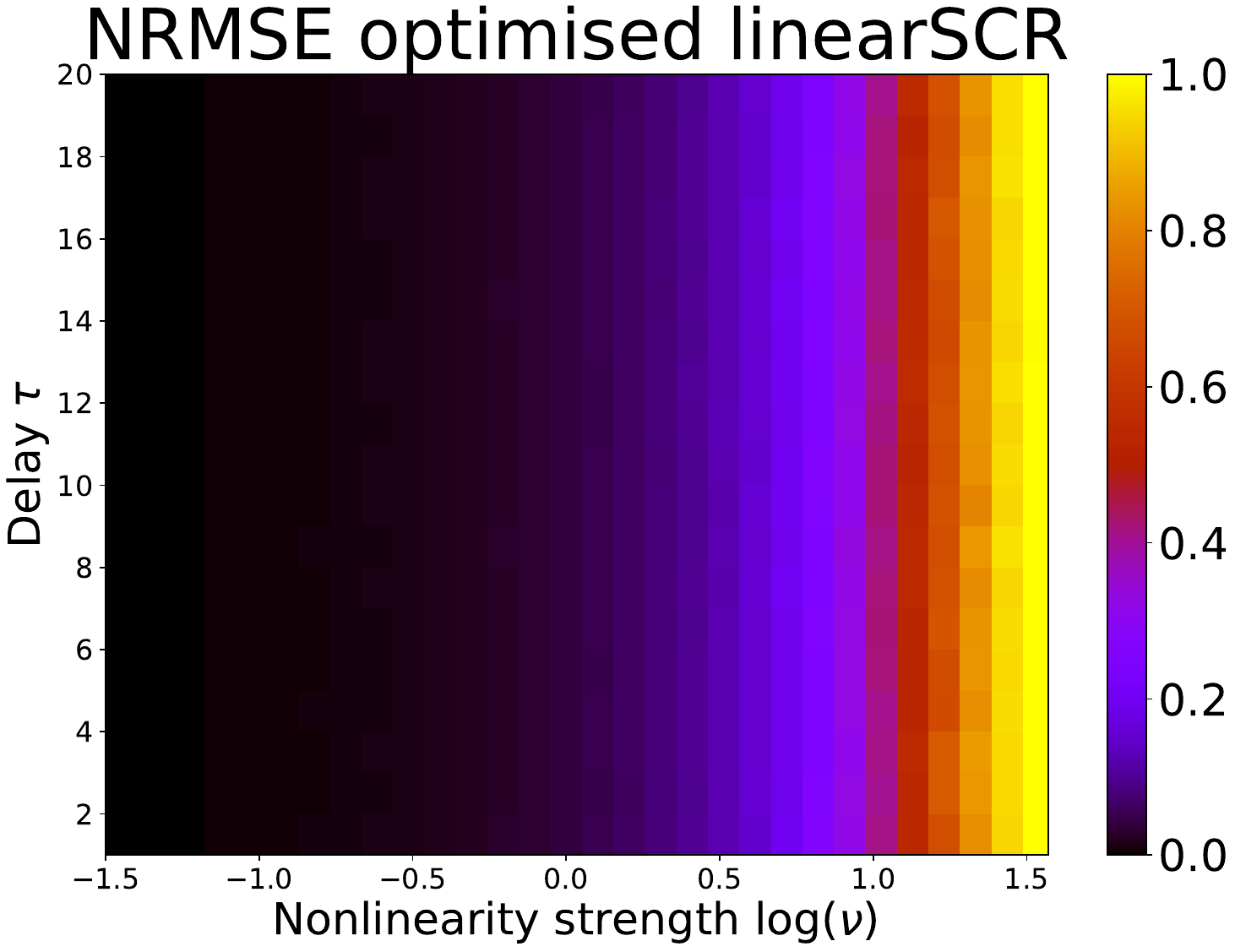}~\includegraphics[keepaspectratio=true,scale=0.21]{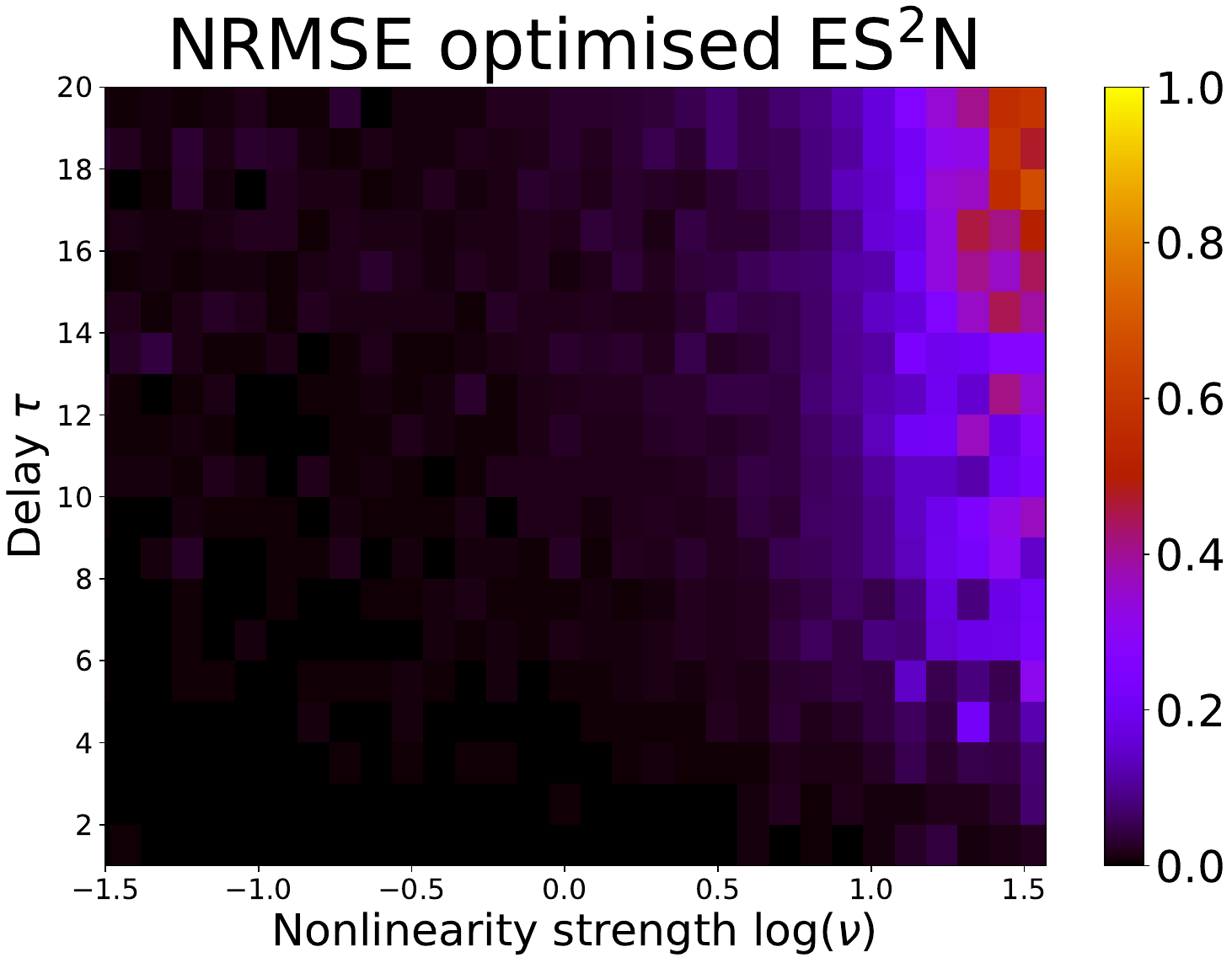}
\caption{Results of the memory-nonlinearity trade-off task explained in Section \ref{sec:inubushi}. NRMSE values ranging from black (NRMSE$=0$) to yellow (NRMSE$=1$ or greater) are plotted for various combinations of delay $\tau$ and nonlinearity strength $\nu$. \textbf{Left:} best NRMSE values after a random search of 100 trials on a leaky ESN. \textbf{Centre:} best NRMSE values after a random search of 100 trials on a \linear{}.  \textbf{Right:} best NRMSE values after a random search of 100 trials on our proposed \model{} model.}
\label{fig:inubushi}
\end{figure}

\begin{figure}[ht!]
    \centering
    \includegraphics[keepaspectratio=true,scale=0.31]{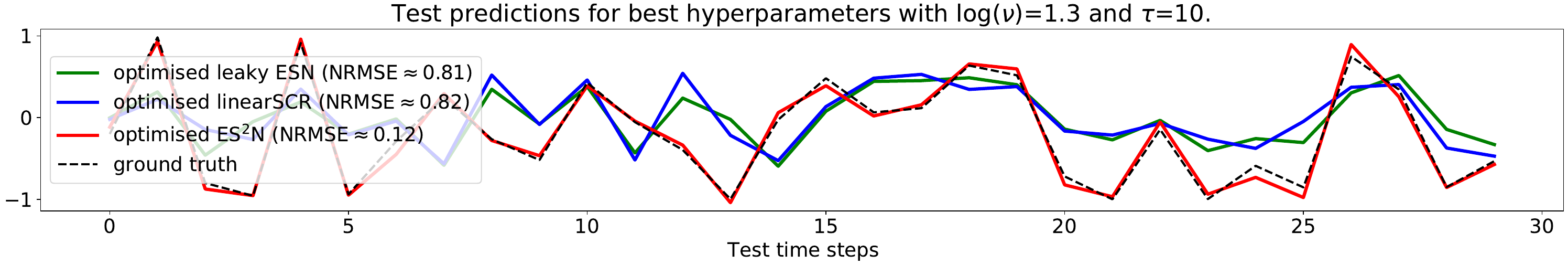}   
\caption{Input $u[t]$ is i.i.d. uniformly sampled in $[-1,1]$. Target is a strong nonlinear transformation of a 10-delayed version of the input given by the function $\sin(\nu u[t-10])$ with $\nu \approx 3.67$ represented in dashed line. The best hyperparameters found for the setting of $\log(\nu)=1.3$ and $\tau=10$ are the following in the form of (input scaling, spectral radius, $\alpha$ or $\beta$): ($0.43, 1.03, 0.98$) for leaky ESN, ($0.35, 0.49, 1.0$) for \linear{}, and ($1.44, 0.10 , 0.18$) for \model{}.}
\label{fig:example_strongnonlinear}
\end{figure}

In the following experiment, we considered a grid of values of $(\log(\nu), \tau)$, with $\tau \in [1,20]$, with a step of $1$, and $\log(\nu) \in [-1.6, 1.6] $ with a step of $0.1$. 
For each pair $(\log(\nu), \tau)$, we ran 100 instantiations of leaky ESN, \linear{}, and \model{}. For each run we generated randomly the following hyperparameters
\begin{itemize}
    \item uniformly random input scaling in $[0.2, 6]$,
    \item uniformly random spectral radii in $[0.1, 3]$,
    \item $\alpha$ (for leaky ESN) and $\beta$ (for \model{}) generated via the formula $a10^{-s}$, with $a$ uniformly random in $(0.1,1)$, and $s$ uniformly random in $\{0,1 \}$, so that they vary in $(10^{-2},1)$.
\end{itemize} 
In Figure \ref{fig:inubushi} are plotted the best NRMSE found on test on a coloured scale from black ($NRMSE=0)$ to yellow ($NRMSE=1$ or greater), the lower the better. 
Results in Figure \ref{fig:inubushi} show that \model{} significantly outperforms both leaky ESN and \linear{}. 
Note that, \linear{} starts to increasingly underperform as soon as $\log(\nu)>0$, i.e. in the region where nonlinearity is needed, while \model{} is able to retrieve the information for much stronger nonlinearly transformed input signals.
This indicates that the \model{} model can truly exploit nonlinearity in the computation. 
On the other hand, leaky ESN is able to reconstruct the input signal in the strong nonlinear regime, but only for very small delays.
%Remarkably, leaky ESN needs a validation session larger than 100 runs to achieve decent results for delays greater than $\tau=7$ time steps. 
In particular, for the challenging case of $\log(\nu)>1$ (strong nonlinearity), leaky ESN's performance significantly degrades already at $\tau=4$, with NRMSE values always above~$ 0.5$. On the contrary, the \model{} model obtains NRMSE values always below of~$ 0.5$ up to delays of $\tau=16$ in the strong nonlinearity regime of $\log(\nu)>1$. \\
In Figure \ref{fig:example_strongnonlinear} are plotted the output signals in the test session for the best hyperparameters found on the three models \model{}, \linear{}, and leaky ESN, in comparison with the ground truth target signal for the case of strong nonlinearity with $\log(\nu) = 1.3$ and $\tau = 10$. 
These results demonstrate how \model{} can conciliate the two contrasting properties of having a large memory capacity and the ability to perform nonlinear computations. 
%This experiment also highlights the little effort needed, in terms of model selection, to find a “good" hyperparameter configuration for the \model{} model, contrary to the case of leaky ESNs.

%The only cases where ESN (with $\alpha=1$) has slightly better performance than ESESN seems to be for $\log(\nu)>1$ and small delays $\tau\leq 3$.
%

\subsection{Multiple superimposed oscillators in auto-regressive mode}
\label{sec:mso}

The Multiple Superimposed Oscillators pattern generation (MSO) is a popular benchmark task where an RNN is trained to generate autonomously (i.e. without the input driving the dynamics) a one-dimensional signal made from the superpositions of a few incommensurate sines.
This is achieved providing in input the target signal during the training phase, then, once the readout matrix is trained, closing the loop in the testing phase via $u[t]=z[t]$ in eq.~\eqref{eq:state_update}, i.e. self-driving the ESN dynamics with its own generated output. In these auto-regressive tasks, the ridge regression training of the readout matrix can be regarded as a \emph{teacher forcing} training strategy.
In the literature \cite{xue2007decoupled, schmidhuber2007training, holzmann2010echo, roeschies2010structure, koryakin2012balanced} the MSO task with different numbers of sine waves has been inspected. In all of the mentioned studies the frequencies of the sine waves were taken from the same set: $\nu_1 = 0.2,
\nu_2 = 0.311, \nu_3 = 0.42, \nu_4 = 0.51, \nu_5 = 0.63, \nu_6 = 0.74, \nu_7 = 0.85$, and $ \nu_8 = 0.97$.  In this section, we only consider the case of 8 frequencies, the most challenging one among those mentioned above, which we will denote concisely with MSO8.
Thus, the target signal takes the following form:
\begin{equation}
    \label{eq:mso}
    y[t] = \sum_{i=1}^{8} \sin(\nu_i t).
\end{equation}
The MSO8 signal of eq.~\eqref{eq:mso} is then rescaled in order to have zero mean and be bounded in $(-1,1)$. From now on, we will refer to $y[t]$ as the normalised signal.
The function of eq.~\eqref{eq:mso} has an almost-period of $\tau_P = 6283$ time steps.\footnote{More precisely, for the normalised signal, it turns out that $ <\bigl|y[t+\tau_P] - y[t]\bigl|>_{t=0, \ldots, \tau_P} \approx 0.024 \pm 0.02 $.}
%Training is accomplished via teacher forcing.
We use 6383 time steps for training, i.e. a whole “period" of 6283 steps excluding the first 100 time steps to wash out transient dynamics. In this phase, the target signal (the teacher) is injected into the reservoir as the input signal, i.e. $u[t] = y[t]$ for $t=1, \ldots, 6383$. During the training phase, a small Gaussian noise of zero mean and standard deviation of $10^{-4}$ has been added to the argument of the $\tanh$ in order stabilise the dynamics. Therefore, the \model{}' state-update equation during training was
\begin{equation}
    \label{eq:noisy_ESESN}
    x[t] = \beta  \phi(\mathbf{W}_{r} x[t-1] +  \mathbf{W}_{in}u[t] + \eta[t]) + (1-\beta) \mathbf{O} x[t-1],
\end{equation}
with $\eta \in N(0, 10^{-4})$ the source of Gaussian noise.
We set zero regularisation, i.e. $\mu=0$ in eq.~\eqref{eq:ridge_regress}. Thus, we train the linear readout to reproduce the target signal. Then, we close the loop feeding back the output into the reservoir in place of the external input, i.e. $u[t]=z[t]$ for $t\geq 6384$. From this moment on, the RNN runs autonomously (without noise).

\subsubsection{Random hyperparameter search}
\label{sec:randomsearch}
First of all, we accomplished a random search to optimise the hyperparameters of both leaky ESN and \model{}. 
Varying the reservoir size revealed that, as expected, larger models leads to better results, for both leaky ESNs and \model{}s.
We considered leaky ESNs of 600 reservoir neurons, and \model{}s of 100 reservoir neurons. We considered leaky ESNs with reservoir size 6 times larger of the \model{}'s, in order for the leaky ESN to get competitive performance on the challenging MSO8 task.
We ran 10000 different initialisations for both leaky ESN and \model{} with uniformly random generated hyperparameters as follows:
\begin{itemize}
    \item uniformly random spectral radius $ \rho \in [0.8, 1.2]$
    \item uniformly random input scaling $\omega_{in} \in [0, 0.4] $
    \item (for leaky ESN) uniformly random $\alpha \in (0.1, 1)$ 
    \item (for \model{}) uniformly random $\beta \in (0.01, 0.1) $.
\end{itemize}
We also tried to vary the reservoir connectivity, but it did not influence the performance. Thus, we used fully connected reservoirs. 
In the training phase, a small Gaussian noise with zero mean and standard deviation of $10^{-4}$ has been introduced in the state-update equation as in eq.~\eqref{eq:noisy_ESESN}, for both leaky ESN and \model{}.
To evaluate the performance we compute the NRMSE as defined in eq.~\eqref{eq:nrmse} for 300 time steps in the testing phase, i.e. for $t=6384, \ldots, 6684$.
We refer the reader to Appendix \ref{sec:results_randomsearch} for the visualisation of the data obtained from this random search. 
In summary, from this search emerged that leaky ESNs need spectral radii strictly around 1 for good results on the MSO8 task. On the contrary, \model{}s are quite insensitive on the choice of the spectral radius. However, by design \model{} reflects the insensitivity on the spectral radius on a dependence on the proximity hyperparameter $\beta$, which for this task reaches its optimum around $\beta=0.03$. Apart from that, the most influential hyperparameter turned out to be the input scaling, in line with previous works. In particular, we found the optimal combination of $\rho =0.99$, $\omega_{in}=0.05$, and $\alpha=0.9$ for leaky ESNs. While for \model{} $\omega_{in}=0.11$, and $\beta =0.03$ (regardless of $\rho$).
All the NRMSEs computed in this random search are reported in the histograms in the left plots of Figure \ref{fig:mso_longrun}. 
The mean and standard deviation of NRMSE for leaky ESN is $1.44\pm 13.34 $, while for the \model{} is $ 0.05 \pm 0.11  $.
The difference of mean NRMSE values highlights the supremacy of \model{} over the leaky ESN on the MSO8 task, despite a 6 times smaller reservoir size.
As a side result, this random search revealed how \model{}s are characterised by a wider ``good'' hyperparameter region compared to leaky ESNs, testified via the more than 120 times larger standard deviation of leaky ESN over \model{}.\footnote{There are few hundreds NRMSE values of leaky ESN greater than 2 which do not appear in the histogram in Figure \ref{fig:mso_longrun}, some of them exceeding NRMSE of 100.}

\subsubsection{Stability in the long run}

\begin{figure}[ht!]
\begin{minipage}{0.1\textwidth}
\centering
    \includegraphics[keepaspectratio=true,scale=0.09]{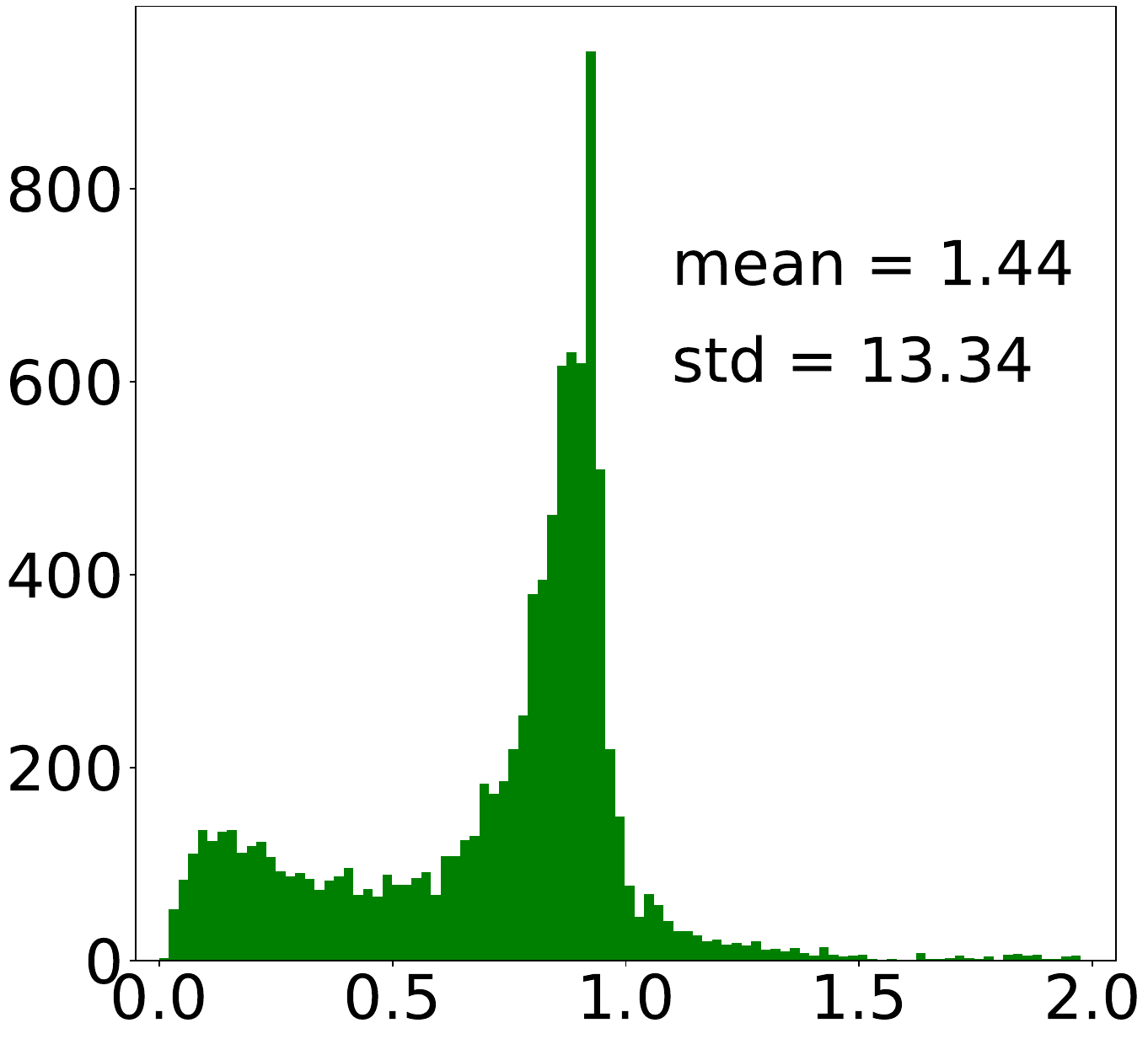}\\
    \includegraphics[keepaspectratio=true,scale=0.09]{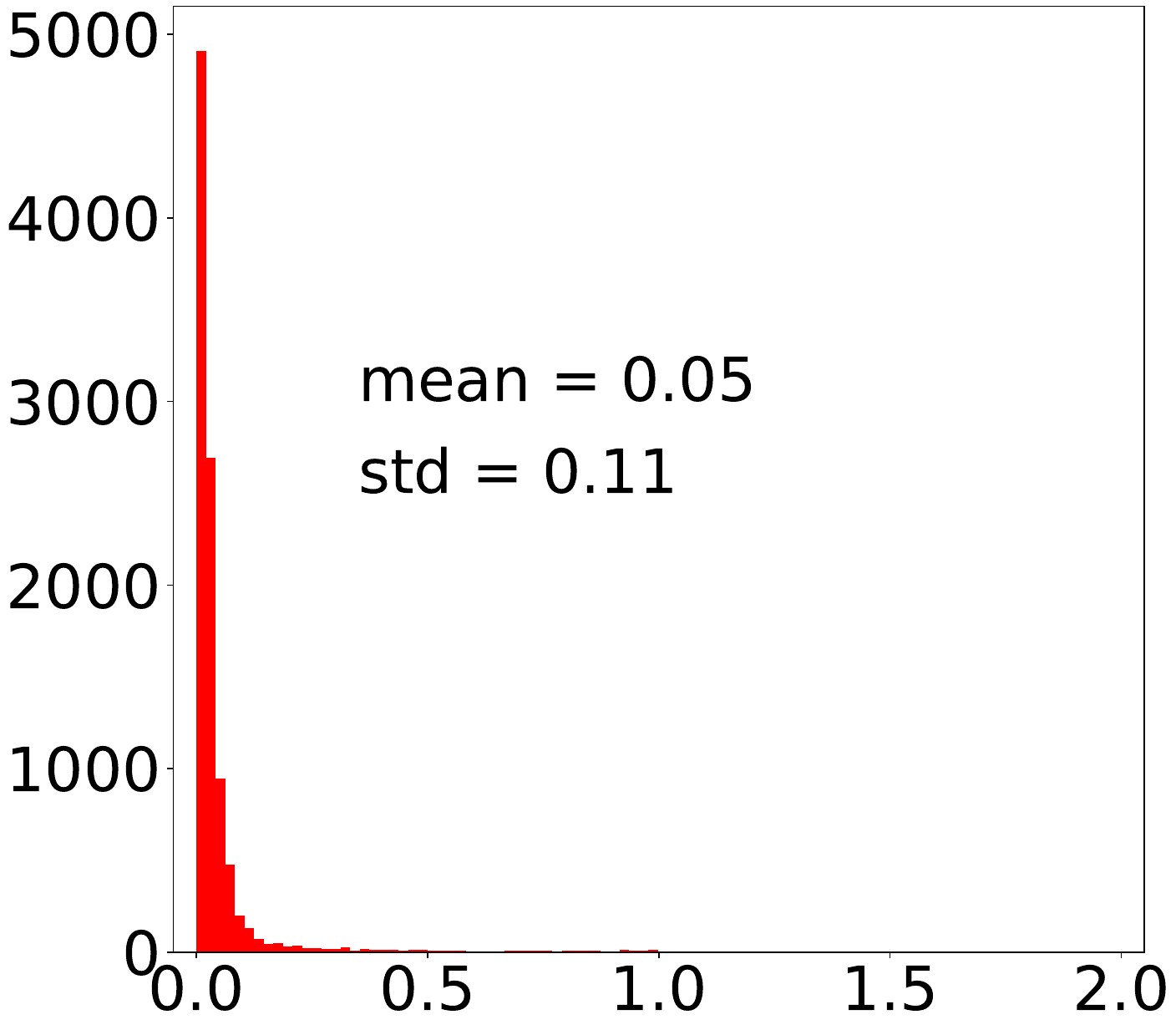}\\
    \includegraphics[keepaspectratio=true,scale=0.09]{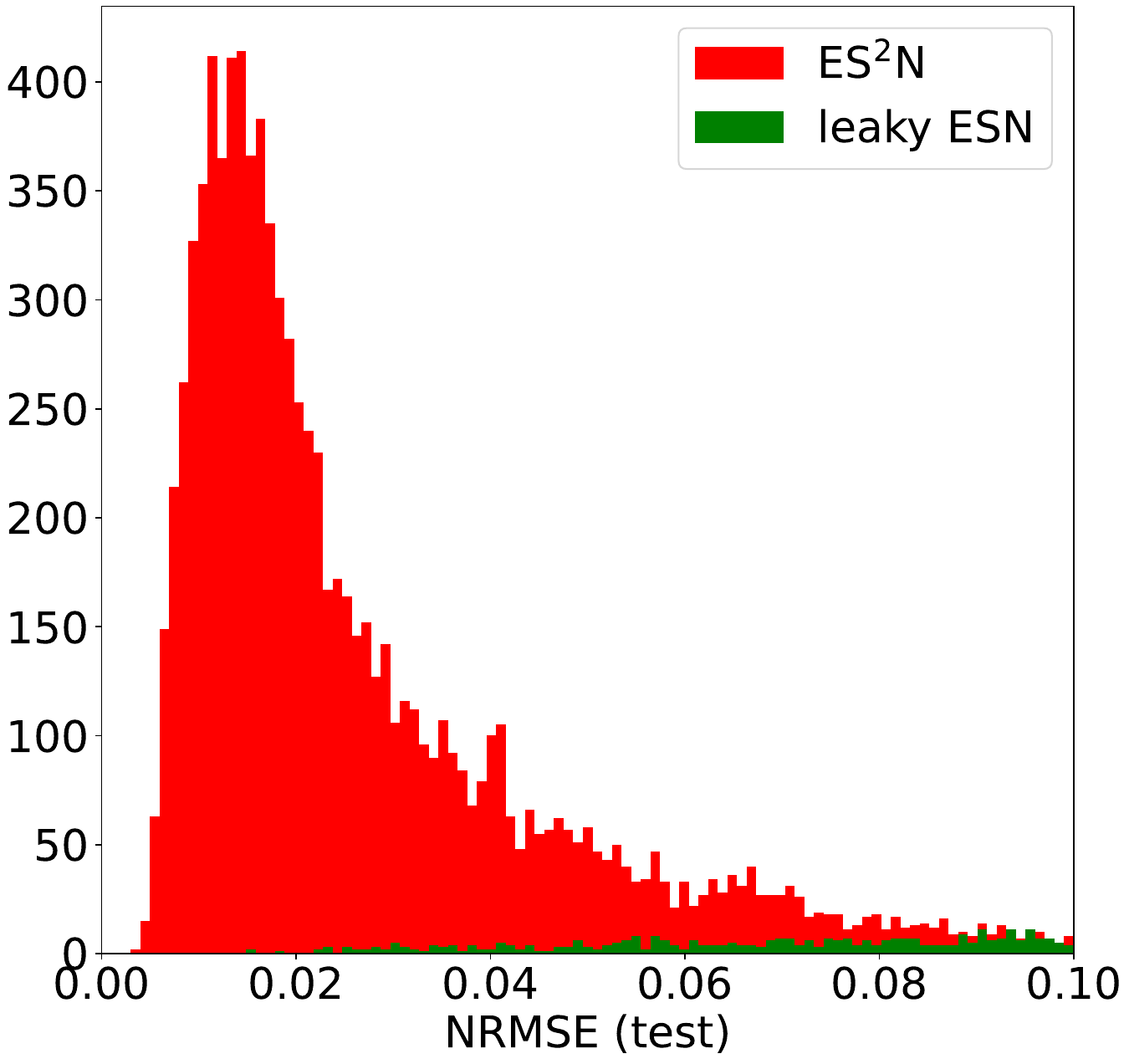}
\end{minipage}~\hspace{0.5cm}
\begin{minipage}{0.65\textwidth}
\centering
    \includegraphics[keepaspectratio=true,scale=0.15]{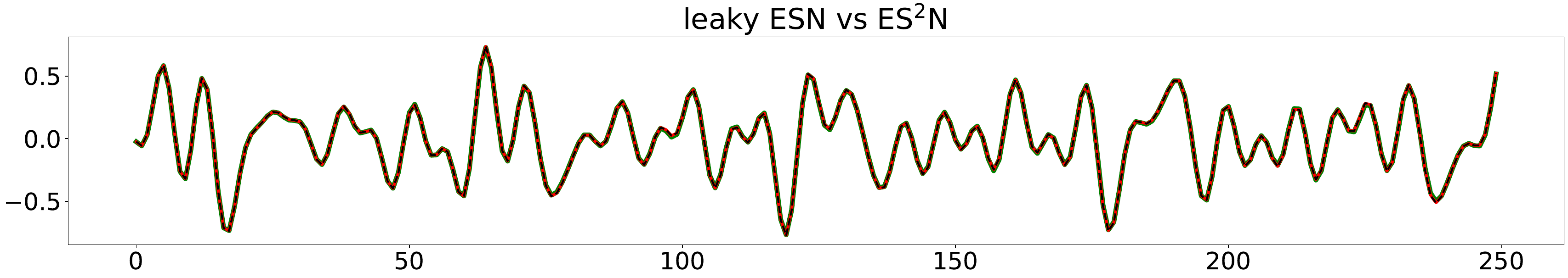}\\
    \includegraphics[keepaspectratio=true,scale=0.15]{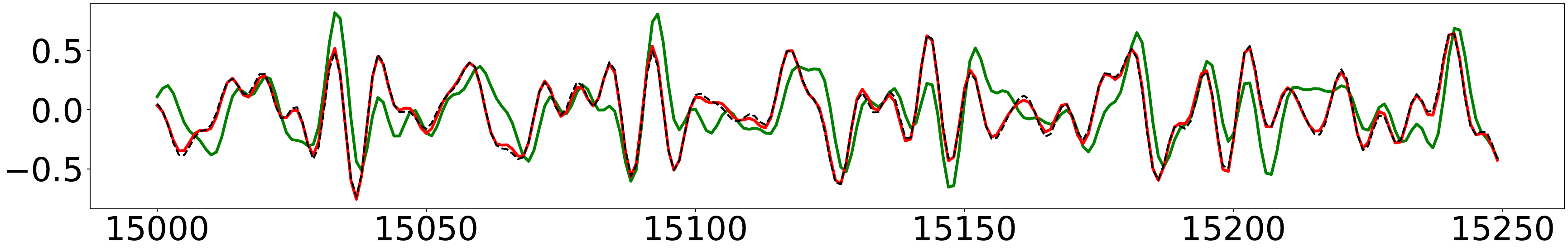}\\   
    \includegraphics[keepaspectratio=true,scale=0.15]{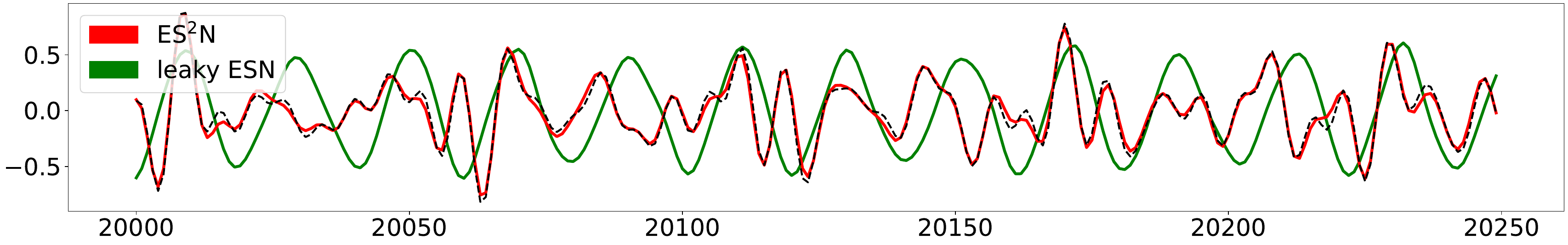}\\ 
    \includegraphics[keepaspectratio=true,scale=0.15]{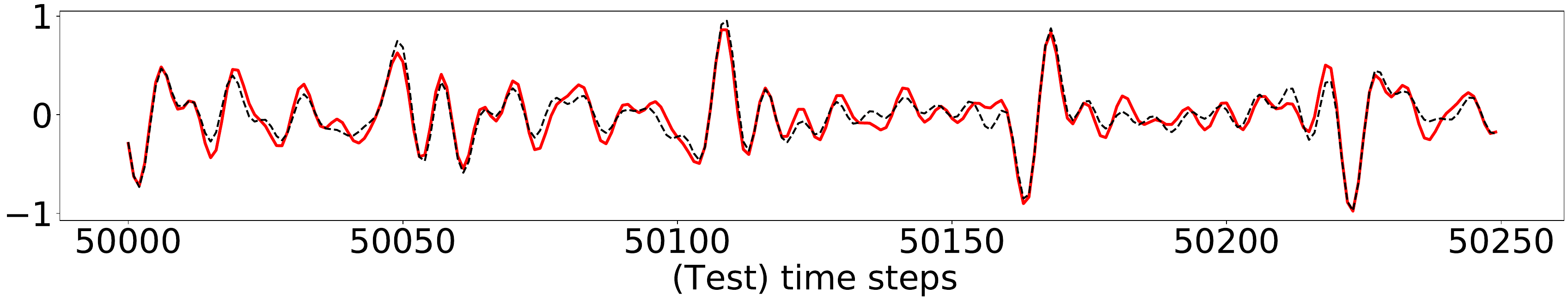}
\end{minipage}~
\begin{minipage}{0.2\textwidth}
\centering
    \includegraphics[keepaspectratio=true,scale=0.13]{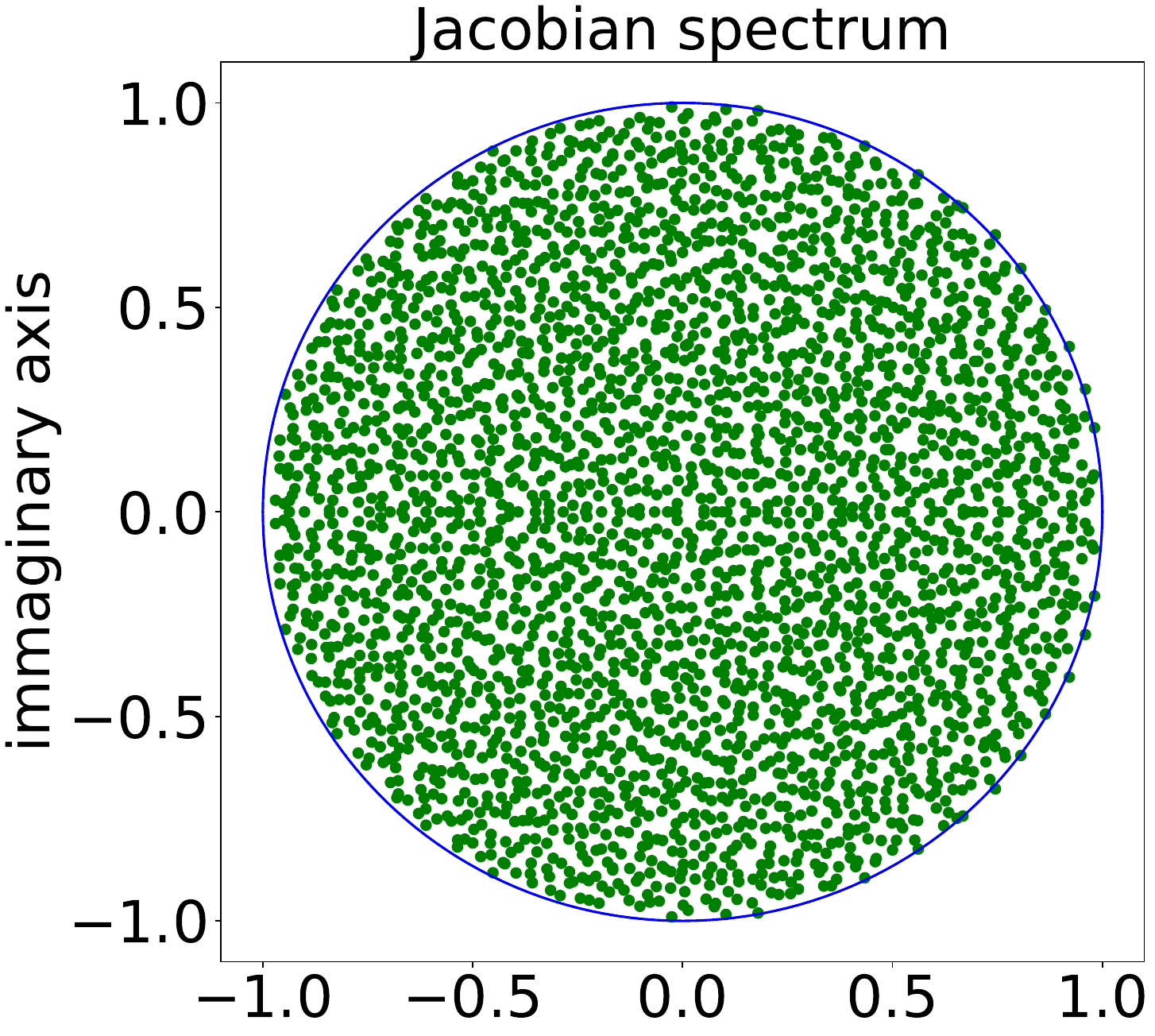}\\
    \includegraphics[keepaspectratio=true,scale=0.13]{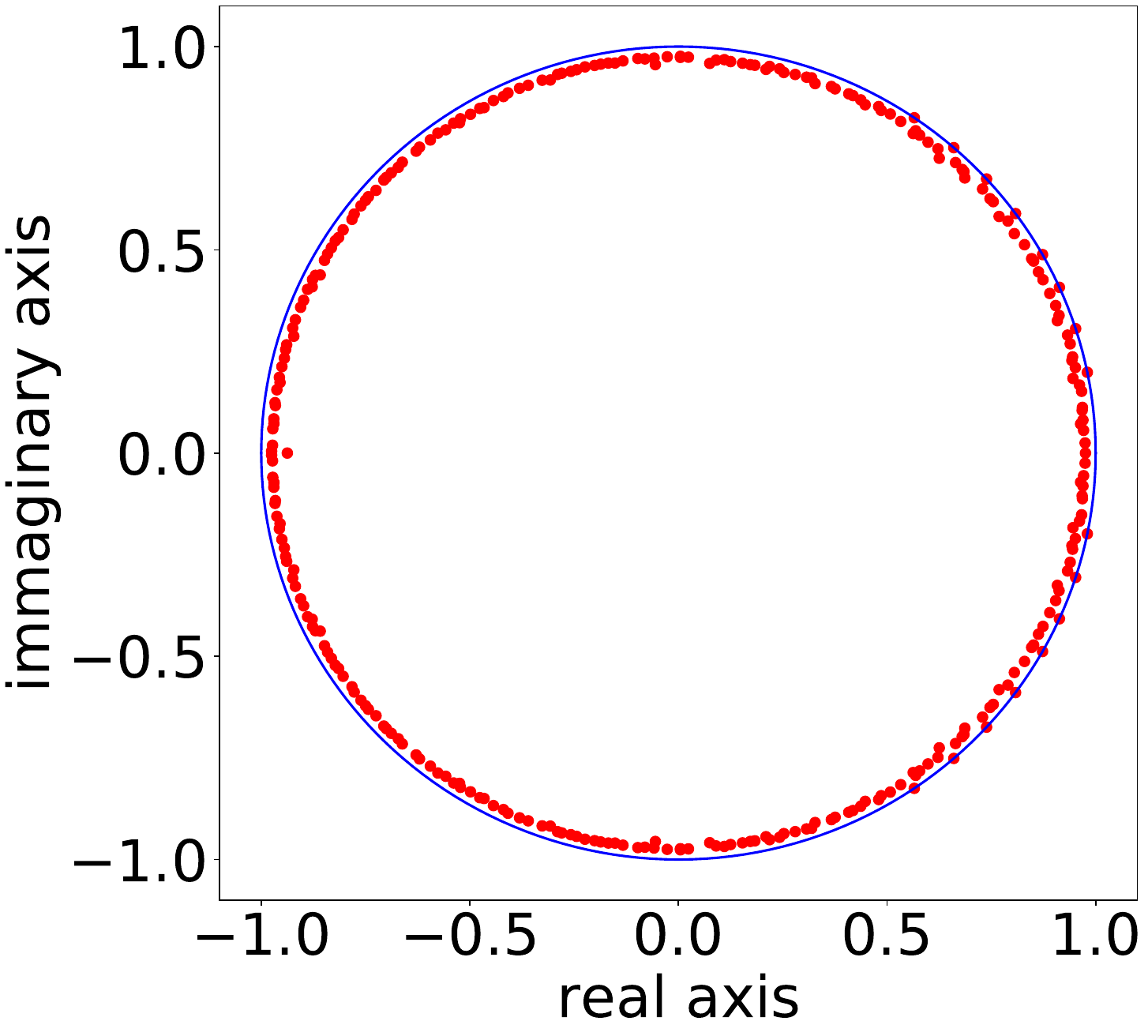}
\end{minipage}
\caption{MSO task with 8 frequencies. In all plots the colour green corresponds to the leaky ESN model, while red to the \model{} model. \textbf{Left:} hystograms of NRMSE of 10000 initialisations of leaky ESN (600 reservoir neurons) and \model{} (100 reservoir neurons) with hyperparameters uniformly generated as explained in Section \ref{sec:randomsearch}. In the upper plot, there are almost 300 cases with NRMSE over 2 which have been cut off from the picture (the maximum reaching a NRMSE of 942). In the bottom plot both leaky ESN and \model{}'s histograms are plotted together and magnified around low NRMSE values of $0.05$. \textbf{Centre and right:} A large fine tuned leaky ESN of 3000 neurons and a relatively small \model{} of 300 neurons have been trained to reproduce with the feedback of the output the MSO8 signal (dashed line). The output signals generated by leaky ESN (green) and \model{} (red) after various time intervals $\Delta t $ of running in auto-generation mode are plotted, from top to bottom, $\Delta t = 0, 15000, 20000, 50000$. On the right plots, the eigenvalues of the Jacobians of the corresponding trained leaky ESN model, and \model{} model, on a randomly selected time step in the testing session.}
\label{fig:mso_longrun}
\end{figure}

In this section, we compare the quality of the learned signal in the long run.
For the purpose, we fixed the hyperparameters in their respective optimal setting, precisely we set $( \rho=1, \omega_{in}=0.11, \beta =0.03 )$ for \model{}s, and $( \rho=0.99, \omega_{in}=0.05, \alpha=0.9 )$ for leaky ESNs.\footnote{The choice of $  \rho = 1 $ for \model{} was arbitrary since it does not influence the outcome of this experiment, see also Appendix \ref{sec:results_randomsearch}.}
Therefore, we trained a large leaky ESN of 3000 neurons, and a relatively small \model{} of 300 neurons, on 6383 training steps as explained above (i.e. 100 for transient, and 6283 for actual training).
The resulting output signals in the test session of these trained models are reported in the centre plots of Figure \ref{fig:mso_longrun}. In the beginning both models follow tightly the target (dashed line). After 15000 time steps the leaky ESN output (green) already deviates significantly from the target, while the \model{} output (red) still performs very well. After 20000 time steps the leaky ESN output is completely decorrelated with the target. On the contrary, the \model{} manages to generate a meaningful output signal even after 50000 time steps of autonomous run. Remarkably, the \model{} model is able to substantially outperform the leaky ESN model while having a number of neurons which is an order of magnitude smaller.

As explained in \cite{jaeger2007optimization}, the MSO is a relatively easy task for a linear ESN, as long as one has at least two reservoir neurons per frequency.
However, the apparent perfection of trained linear ESNs hides an intrinsic unstable phase-coupling dynamics, which manifests itself as soon as little perturbations are applied to the system.
On the other hand, nonlinear ESNs are more resilient to perturbations, but they struggle to learn functions composed of even just two superimposed oscillators, especially if one wishes to maintain the learned oscillations for long time.
The \model{} is a nonlinear model lying in between these two extremes, since it can be trained to self-sustain complex oscillatory dynamics for very long time, as shown in the centre plots of Figure \ref{fig:mso_longrun}. This can be attributed to the peculiar form of the \model{}'s Jacobian spectrum which tends to dispose its eigenvalues along the unitary circle promoting the dynamics to take place at the edge of stability as evident from the right plots in Figure \ref{fig:mso_longrun}.

\section{Conclusions} 
\label{sec:conclusions}

In this paper, we have developed a new RC architecture known as the Edge of Stability Echo State Network (\model{}), which has the unique feature of being controllably tunable to the edge of chaos dynamical behavior. 
%is obtained as a convex combination of a non-linear input-driven component, and a linear input-free orthogonal one. 
%takes advantage of both the benefits of having a linear orthogonal reservoir and a nonlinear contracting dynamics. 
Our mathematical analysis first showed that the proposed model has dynamic behavior that is able to exhibit contracting dynamics and the Echo State Property in a manner similar to a standard ESN.
Furthermore, and relevantly, we provided precise analytical bounds for the entire eigenspectrum of the Jacobian of the forward map that hold on each input-driven trajectory and for all time steps. As a result, the reservoir exhibits a behavior whose quality is determined by a specific proximity hyper-parameter. By architectural construction, smaller values of this hyper-parameter result in dynamics that are progressively closer to the edge of chaos. 
Overall, the introduced model takes advantage of both the benefits of having a linear orthogonal reservoir and a nonlinear contracting dynamics.

%We analysed the conditions for the echo state property to hold, and we provided precise analytical bounds for the entire eigenspectrum of the Jacobian of the forward map that hold on each input-driven trajectory and for all time steps.

%Unfolding the recurrent equation of the linear version of \model{} reveals how our proposed model takes advantage of both the benefits of having a linear orthogonal reservoir and a nonlinear contracting dynamics. 

We empirically showed that \model{} can reach the maximum memory capacity obtainable within a given reservoir size, showing significant advantages over standard nonlinear and linear ESN alternatives.
%an impossible feature to get from linear models.
Furthermore, we tested the trade-off between nonlinear computation and long short-term memory, and found that \model{} can reconstruct strongly nonlinear transformations of relatively large delayed input signals, where both standard (nonlinear) ESN and linear ESN fail.
Finally, we empirically demonstrated the superiority of \model{} in the generation of complex oscillatory patterns. Remarkably, the recurrent network driven with its own output signal can produce meaningful oscillations for 50 thousands time steps of autonomous run on the MSO8 task.

The analytical and experimental results presented in this paper have shown, already in this form, the advantages of combining nonlinear dynamics and orthogonal transformations in the state space of a recurrent network. 
Building upon these findings, our future research aims to investigate alternative forms of reservoir construction in a \model{}. For instance, we plan to explore reservoir construction methods based on permutation or circular shift matrices, which on the one hand allow to further improve the computational efficiency of the approach while maintaining its computational properties, and on the other hand are prone to implementation in edge or neuromorphic devices. 
Moreover, in forthcoming studies, we will explore the performance of \model{} in various applications, such as time series forecasting, attractor reconstruction, and classification tasks.
%Furthermore, from an application point of view, in future works we intend to explore the performance of \model{} in forecasting of time series, reconstruction of attractors, and classification tasks.
%Regarding the first two, preliminary results suggest that optimal \model{}s are often characterised by $\beta$ values close to 1, thus leaning to coincide with the leaky ESN model.
%A more in depth analysis is needed to understand the reason behind this behaviour. However, it has already been pointed out in the literature \cite{marzen2017difference} that optimising memory capacity does not imply good forecasting capabilities, hence suggesting the possibility that \model{}'s performance might at most match those of a leaky ESN in these tasks.
%While regarding classification tasks, we expect to obtain good results due to the high memory capacity of the \model{} model, especially for time series characterised by long-term dependencies.

\section*{Acknowledgements}
This work is partially supported by the EC H2020 programme under project TEACHING (grant n. 871385), and by the EU Horizon research and innovation programme under project EMERGE (grant n. 101070918).

\bibliographystyle{abbrvnat}
\bibliography{biblio.bib}

%%%%%%%%
\clearpage
\appendices

\iffalse
%##### ESESN with SimpleCycle #####
\section{\model{} with simple cycles}
\label{sec:esesn_simplecycle}

In this supplemental experiment we implemented a circular shift as orthogonal matrix $\mathbf{O}$ in eq.~\eqref{eq:state_update} with different lengths of the cycle.
We computed the memory capacity for various cycle lengths varying $\beta$, and plotted the $MC_k$ for various delays $k$ (for optimal $\beta$ values found), plotted respectively in Figure \ref{fig:MC_various_cycles} left and right.
The cycle length $L$ means that the orthogonal matrix is formed by two diagonal blocks: the first block is the $L\times L $ orthogonal matrix representing a simple cycle of length $L$, and the second block is the identity matrix of dimension $(100-L)\times (100-L)$.
%
\begin{figure}
    \centering
    \includegraphics[keepaspectratio=true,scale=0.35]{figure/MC_varying_beta_variousCycles.pdf}~\hspace{0.5cm}\includegraphics[keepaspectratio=true,scale=0.35]{figure/MCk_various_cycles.pdf}
\caption{\textbf{Left:} MC curves varying $\beta$ for various cycle lenghts. \textbf{Right:} $MC_k$ values for various cycle lengths, computed for the optimal $\beta$ value found.}%
\label{fig:MC_various_cycles}
\end{figure}

Interestingly, implementing \model{} with circular shifts as orthogonal matrices we can precisely design the amount of short-term memory needed from our model via tuning the length of the cycle. 
%##### ESESN with SimpleCycle #####
\fi

\section{Data from the random search of the MSO8 task}
\label{sec:results_randomsearch}

Here we report the NRMSE values of the random search of the 10000 runs for the MSO8 task for both the \model{} model (left plots), and the leaky ESN model (right plots).
Note that a great portion of the 10000 runs of the \model{} model has NRMSE value less than 2e-2, while the vast majority of runs of the leaky ESN model has NRMSE value greater than 1e-1.
\begin{figure}[ht!]
    \centering
    \includegraphics[keepaspectratio=true,scale=0.14]{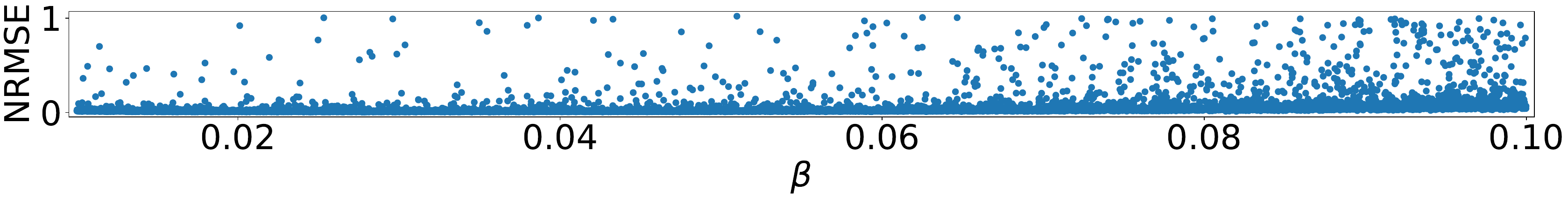}~\hspace{0.5cm}~
    \includegraphics[keepaspectratio=true,scale=0.14]{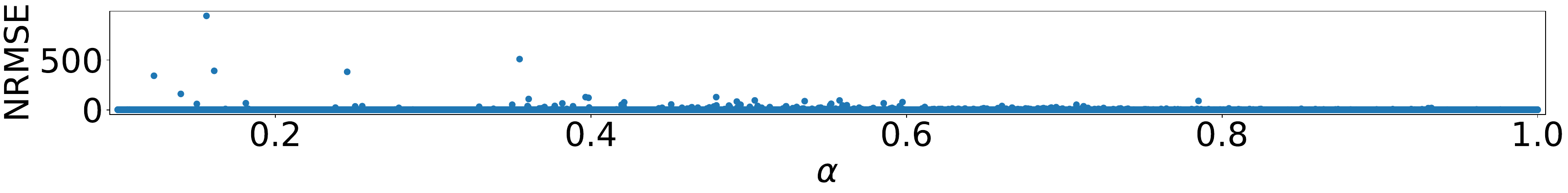}\\
    \includegraphics[keepaspectratio=true,scale=0.14]{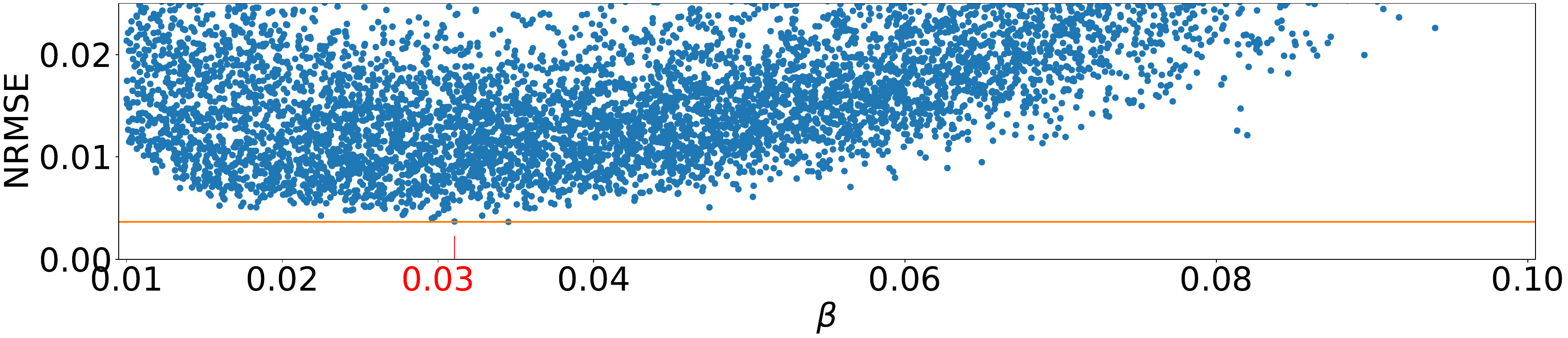}~\hspace{0.3cm}~
    \includegraphics[keepaspectratio=true,scale=0.14]{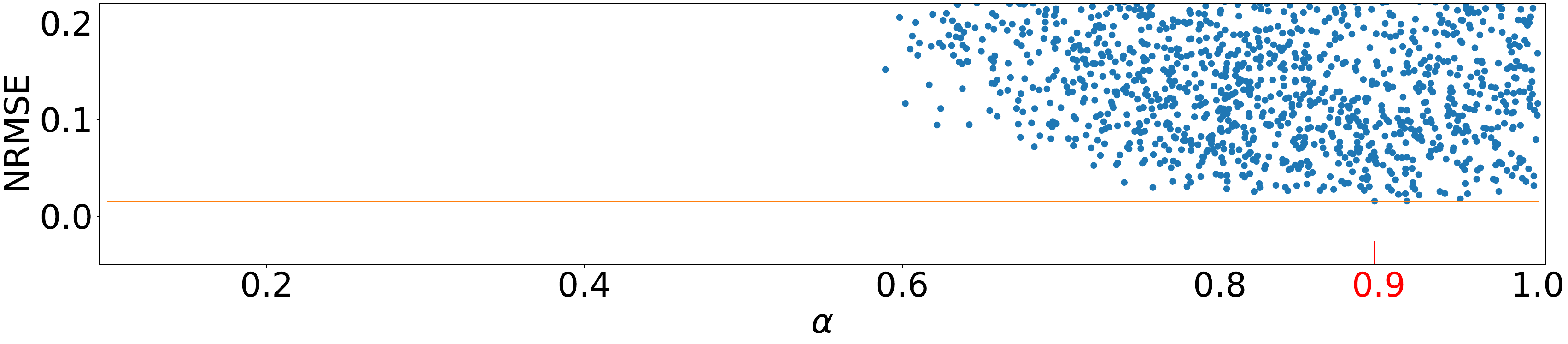}\\
        \vspace{0.9cm}
    \includegraphics[keepaspectratio=true,scale=0.14]{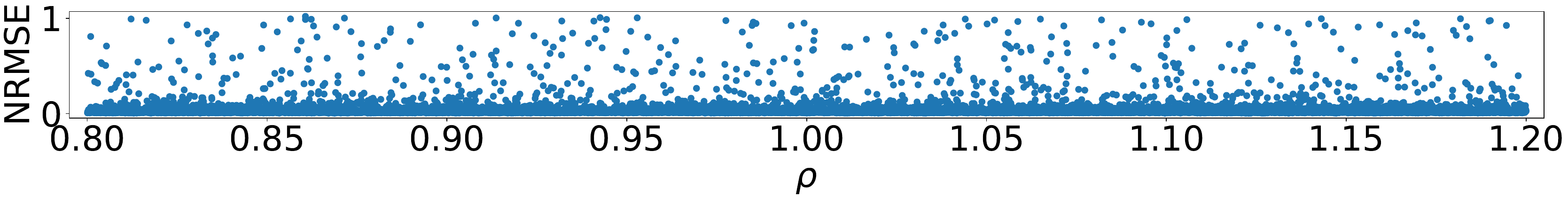}~\hspace{0.5cm}~
    \includegraphics[keepaspectratio=true,scale=0.14]{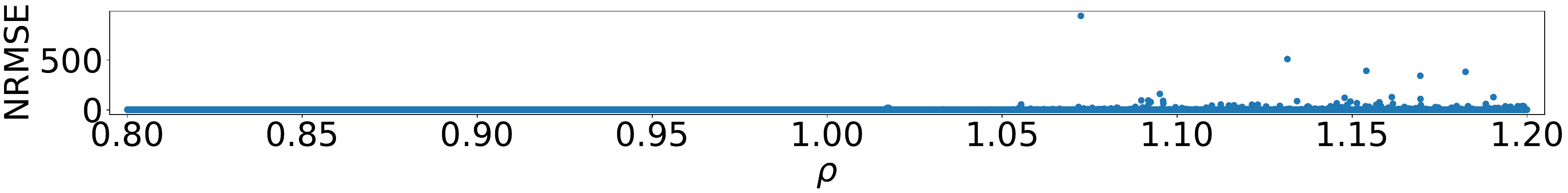}\\
    \includegraphics[keepaspectratio=true,scale=0.14]{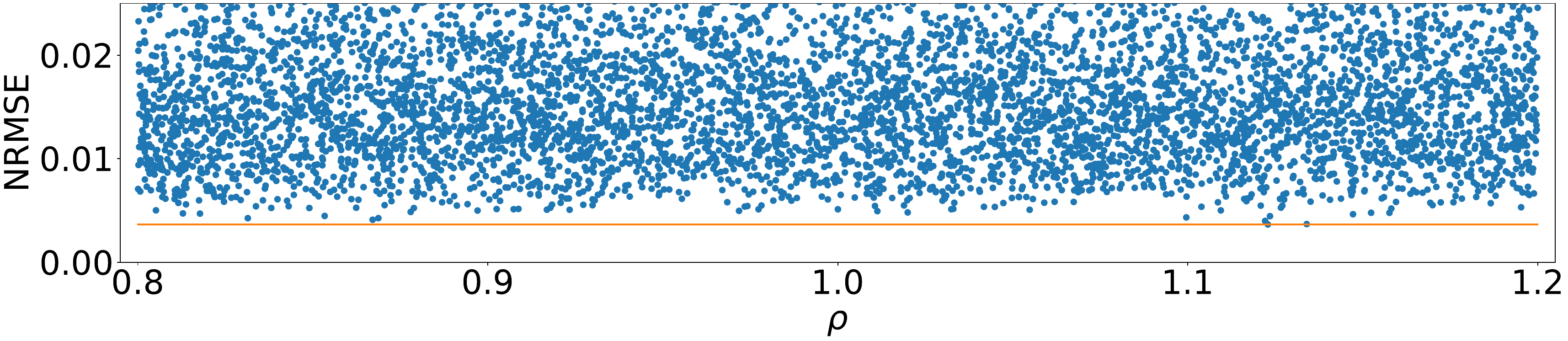}~\hspace{0.3cm}~
    \includegraphics[keepaspectratio=true,scale=0.14]{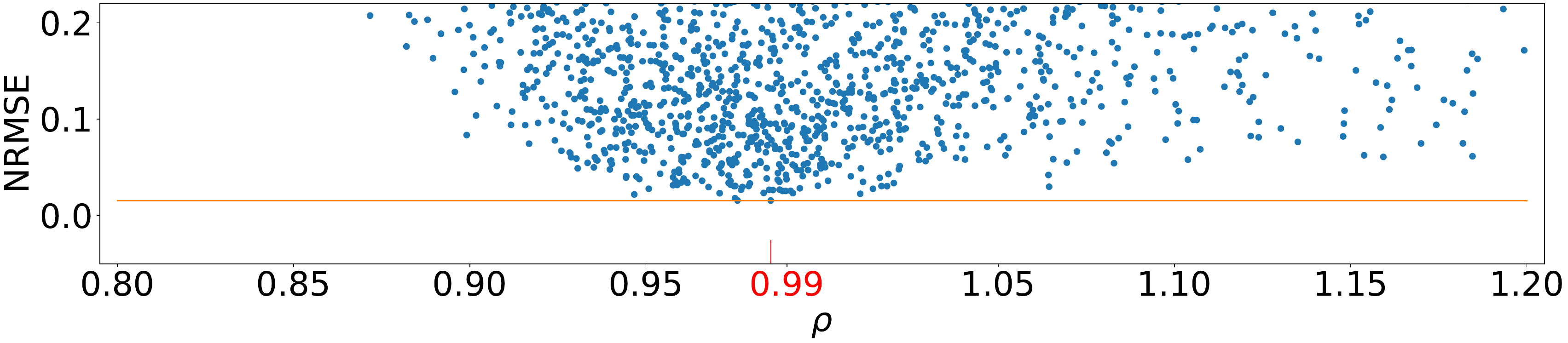}\\
        \vspace{0.9cm}
    \includegraphics[keepaspectratio=true,scale=0.14]{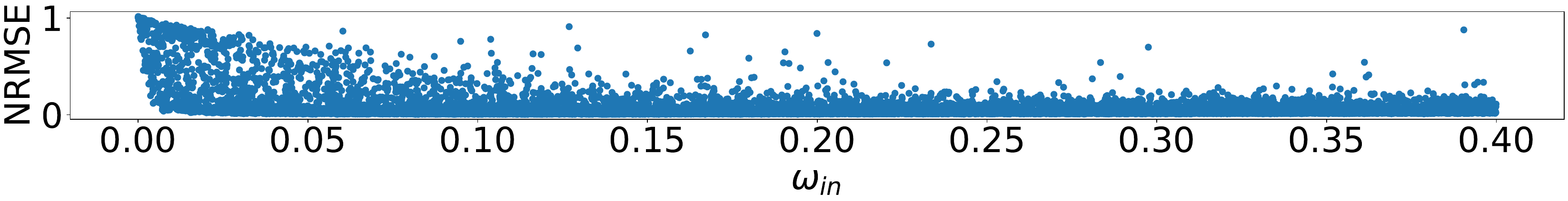}~\hspace{0.5cm}~
    \includegraphics[keepaspectratio=true,scale=0.14]{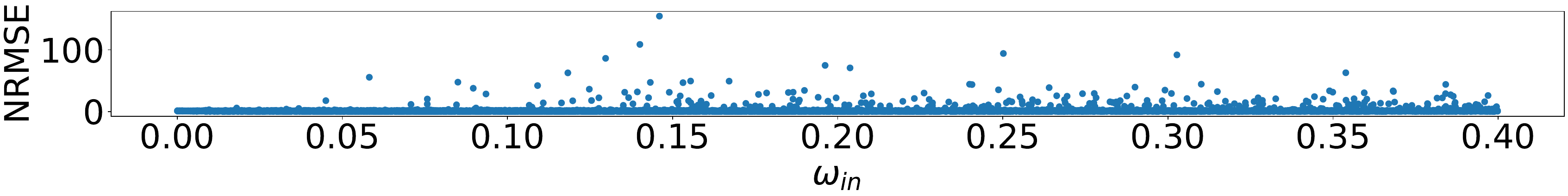}\\
    \includegraphics[keepaspectratio=true,scale=0.14]{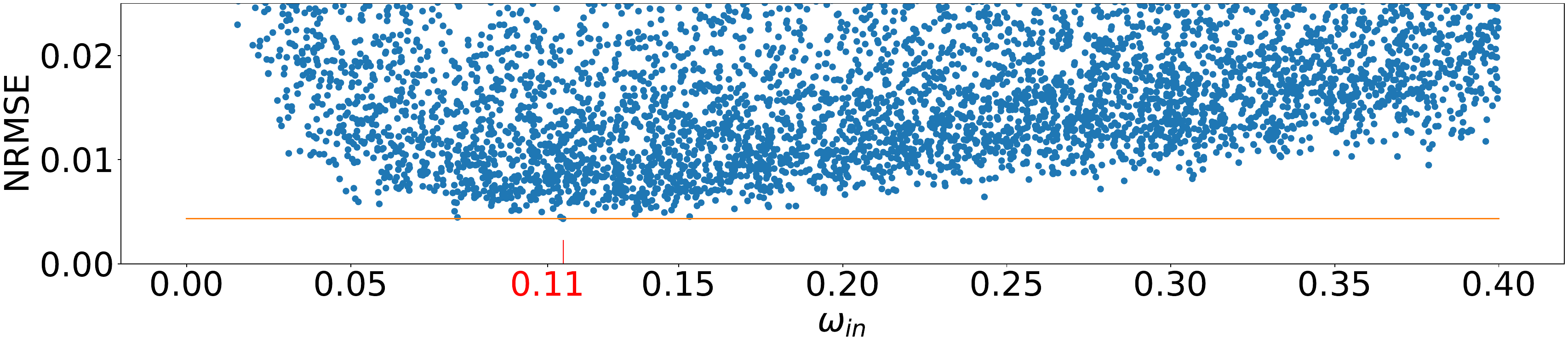}~\hspace{0.3cm}~
    \includegraphics[keepaspectratio=true,scale=0.14]{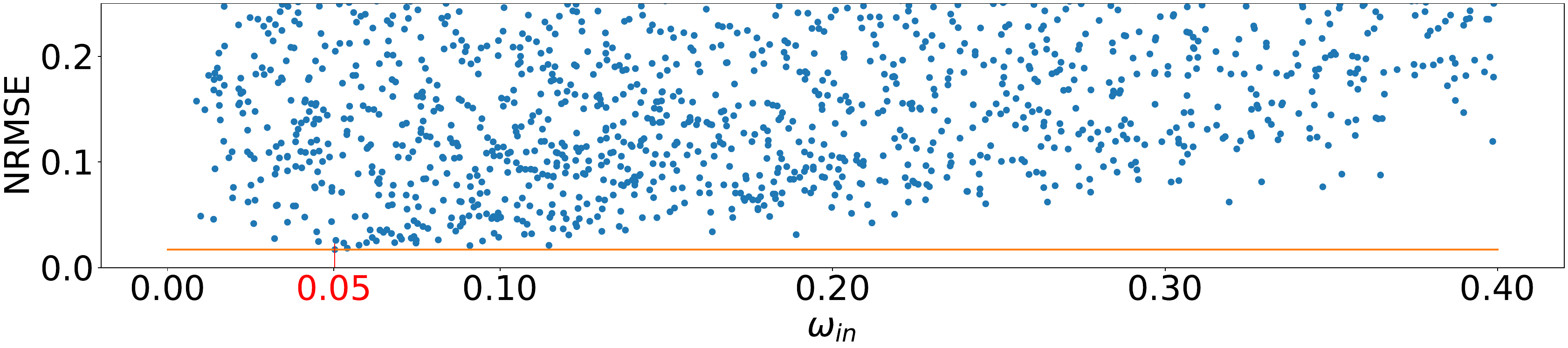}
\caption{NRMSEs resulted from the random search of section \ref{sec:randomsearch} plotted versus each one of the hyperparameter. The horizontal orange lines correspond to the minimum NRMSE values reached, while the corresponding minimum points are highlighted in red on the abscissa. \textbf{Left:} data of \model{}. \textbf{Right:} data of leaky ESN.}
\label{fig:random_search_MSO8}
\end{figure}

\end{document}